\documentclass[10pt,twocolumn,letterpaper]{article}

\usepackage[pagenumbers]{wacv} 

\usepackage{graphicx}
\usepackage{amsmath}
\usepackage{amssymb}
\usepackage{booktabs}
\usepackage{microtype}
\usepackage{subcaption}
\usepackage{multirow} 
\usepackage{amsmath}
\usepackage{amssymb}
\usepackage{mathtools}
\usepackage{amsthm}
\usepackage{algorithm}
\usepackage{algpseudocode}
\usepackage[pagebackref,breaklinks,colorlinks]{hyperref}

\usepackage{soul}

\usepackage[capitalize]{cleveref}
\crefname{section}{Sec.}{Secs.}
\Crefname{section}{Section}{Sections}
\Crefname{table}{Table}{Tables}
\crefname{table}{Tab.}{Tabs.}

\def\wacvPaperID{
1038} 
\def\confName{WACV}
\def\confYear{2025}

\title{Make VLM Recognize Visual Hallucination on Cartoon Character Image with Pose Information}

\author{
    Bumsoo Kim$^{1*}$, Wonseop Shin$^{1*}$, Kyuchul Lee$^{2\ddagger}$, Yonghoon Jung$^{1}$, Sanghyun Seo$^{1\dagger}$\\
    $^1$Chung-Ang University, $^2$Coupang\\
    {\tt\small \{$^{1*}$bumsookim, $^{1*}$wonseop218, $^1$dydgns2017, $^{1\dagger}$sanghyun\}@cau.ac.kr, $^{2\ddagger}$kylee287@coupang.com}
}
\begin{document}
\maketitle

{
  \renewcommand{\thefootnote}%
    {\fnsymbol{footnote}}
  \footnotetext[1]{ Equal contributor} 
  \footnotetext[3]{ Project leader}
  \footnotetext[2]{ Corresponding author}
} 

\begin{abstract}
Leveraging large-scale Text-to-Image (TTI) models have become a common technique for generating exemplar or training dataset in the fields of image synthesis, video editing, 3D reconstruction. However, semantic structural visual hallucinations involving perceptually severe defects remain a concern, especially in the domain of non-photorealistic rendering (NPR) such as cartoons and pixelization-style character. To detect these hallucinations in NPR, We propose a novel semantic structural hallucination detection system using Vision-Language Model (VLM). Our approach is to leverage the emerging capability of large language model, in-context learning which denotes that VLM has seen some examples by user for specific downstream task, here hallucination detection. Based on in-context learning, we introduce pose-aware in-context visual learning (PA-ICVL) which improve the overall performance of VLM by further inputting visual data beyond prompts, RGB images and pose information. By incorporating pose guidance, we enable VLMs to make more accurate decisions. Experimental results demonstrate significant improvements in identifying visual hallucinations compared to baseline methods relying solely on RGB images. Within selected two VLMs, GPT-4v, Gemini pro vision, our proposed PA-ICVL improves the hallucination detection with 50\% to 78\%, 57\% to 80\%, respectively. This research advances a capability of TTI models toward real-world applications by mitigating visual hallucinations via in-context visual learning, expanding their potential in non-photorealistic domains. In addition, it showcase how users can boost the downstream-specialized capability of open VLM by harnessing additional conditions. We collect synthetic cartoon-hallucination dataset with TTI models, this dataset and final tuned VLM will be publicly available. The dataset and demo VLMs are provided in the corresponding page: \href{https://gh-bumsookim.github.io/Cartoon-Hallucinations-Detection/}{https://gh-bumsookim.github.io/Cartoon-Hallucinations-Detection/}. 
\end{abstract}

\vspace{-1em}
\section{Introduction}
\label{sec:intro}

Leveraging machine-generated images based on large Text-to-Image (TTI) models \cite{rombach2022high, nichol2021glide, ramesh2021zero, podell2023sdxl} has become a prominent, efficient technique in the field of image synthesis \cite{yang2022pastiche}, video generation \cite{blattmann2023stable}, 3D reconstruction \cite{zhang2023styleavatar3d, voleti2024sv3d}. Recently, beyond TTI generation, complex multi-modal approaches combining Visual Language Models (VLMs) \cite{zhang2023vision, achiam2023gpt, liu2024visual, alayrac2022flamingo, reid2024gemini} upon Large Language Models (LLMs) \cite{zhao2023survey, touvron2023llama} are being utilized together with TTI to obtain more compelling, imaginative and plausible images. This is achieved by using reference images \cite{zhang2023adding} or human-level image captioning \cite{hu2022scaling} with multi-turn prompting under the advantages of in-context visual learning \cite{wang2023images, zhang2024makes}.

\begin{figure}[t]
\centering
    \includegraphics[width=0.9\linewidth]{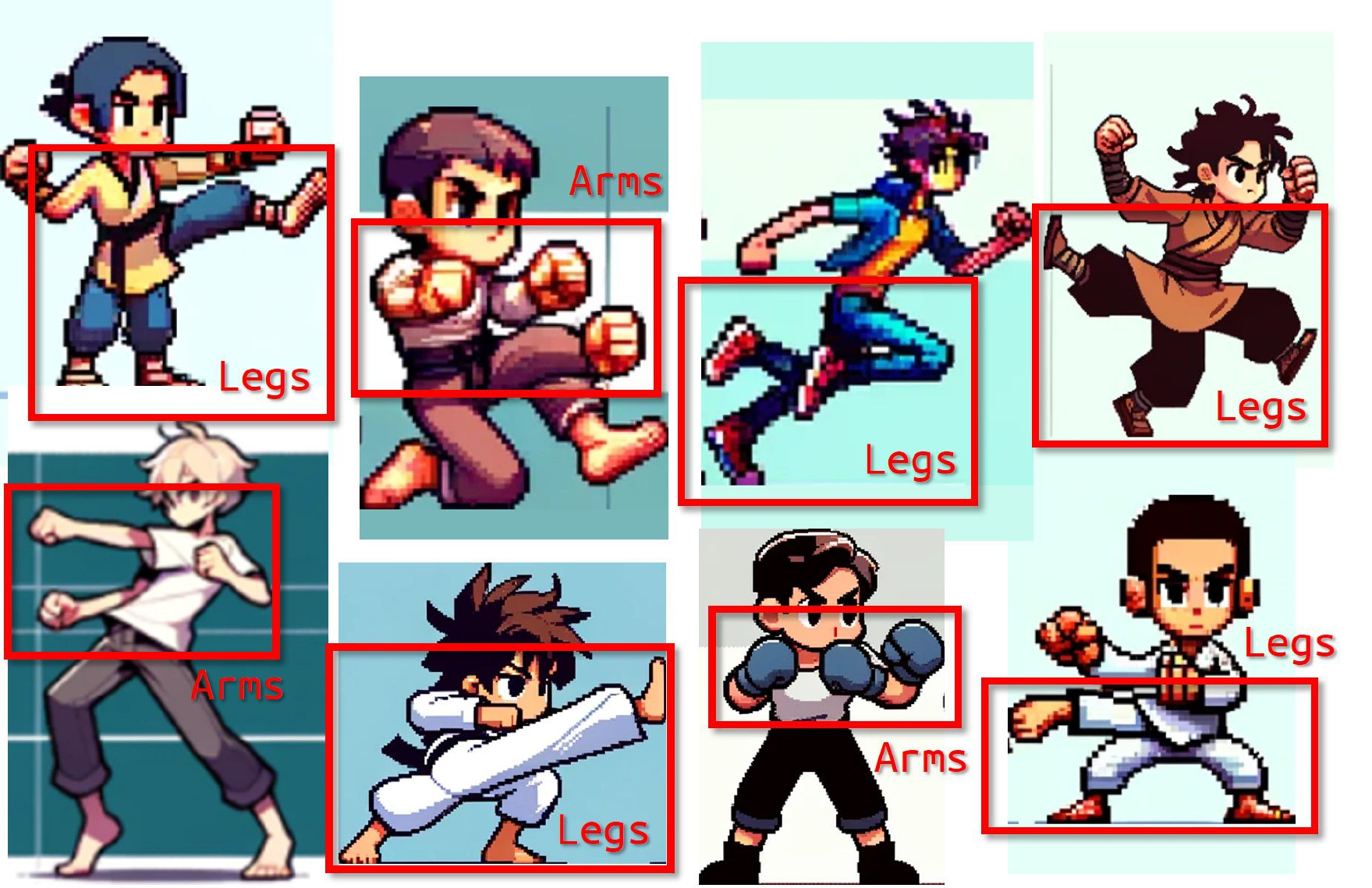}
\caption{Examples of \textit{semantic structural hallucination} in cartoon/pixel rendering images generated by TTI models. These hallucination hinder the TTI model to be extended toward applications, additionally requiring that users embark on burdensome process of eliminating hallucination sample in manual. More samples can be found in Appendices \ref{suppl:cartoon_domain_hallucination}.}
\vspace{-1.0em}
\label{fig:hallucination_example}    
\end{figure}

However, TTI and VLMs \cite{li2023evaluating} often suffer from the visual hallucination problem, similar to the hallucination issue in language models under LLMs \cite{huang2023survey, xu2024hallucination}. In image and video formats, a phenomenon, we denote it as \textit{semantic structural hallucination}, occurs when images appear clear at first glance but show major inaccuracies upon closer examination as shown in Fig. \ref{fig:hallucination_example}. This means that even though the images seem correct initially, they actually contain errors that become obvious when you look more closely. These inaccuracies compromise the models' reliability and trustworthiness, posing challenges for network training and the broader adoption of large-scale generative models. Recent work, Vision language models are blind\cite{rahmanzadehgervi2024vision}, explored the seven BlindTest, a basic visual tasks, and identified that VLM has frequently failed to understand the visual structure with textual prompts. It demonstrated that VLM has a still room to be enhanced toward specific task, remaining several frontier to be explored, even for basic visual mission. 

\begin{figure}[t]
  \centering
  \begin{subfigure}{0.23\textwidth}
    \includegraphics[width=\linewidth]{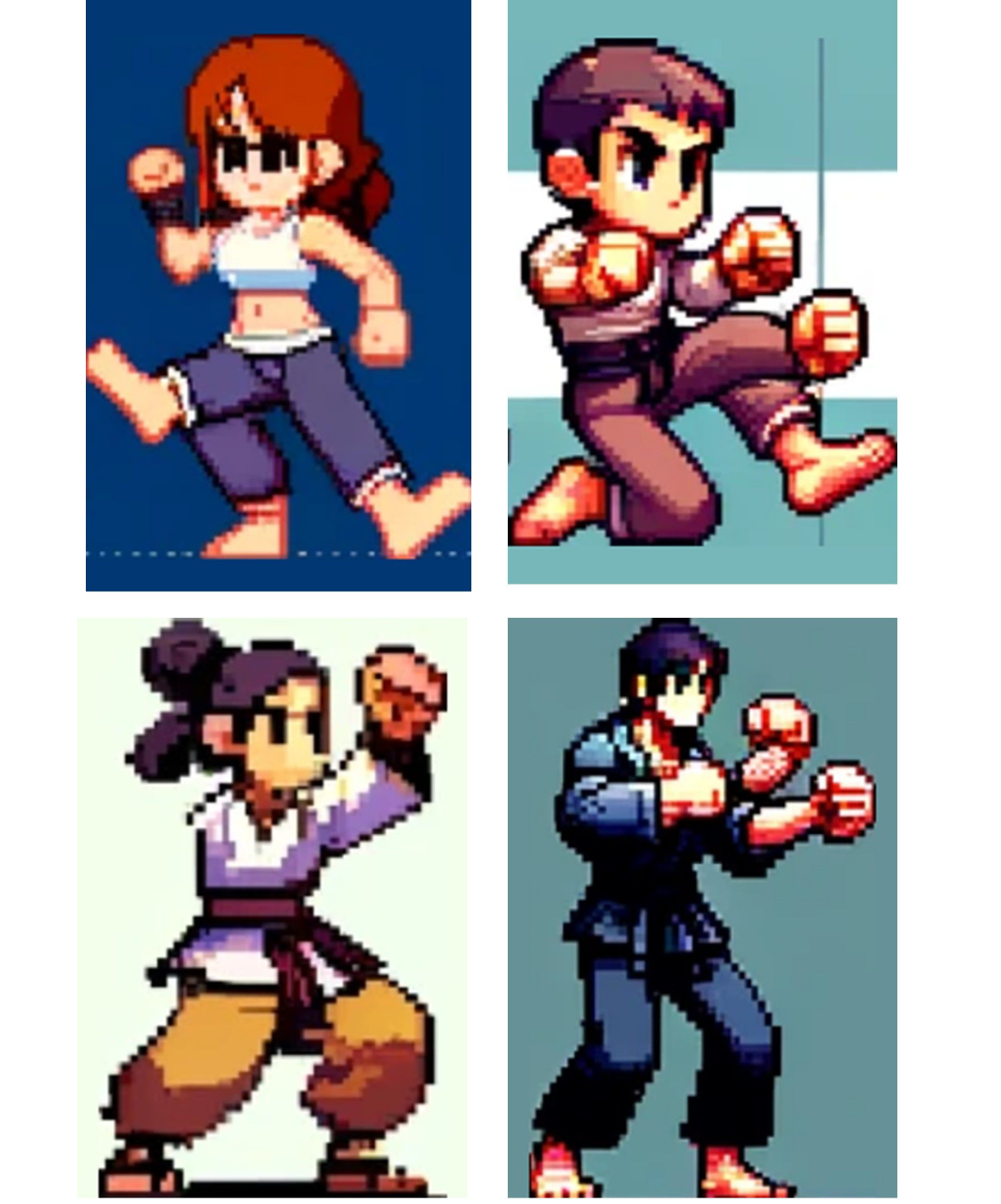}
    \caption{Real hallucination.}
    \label{fig:real_hallucination}
  \end{subfigure}
  \hfill
  \begin{subfigure}{0.23\textwidth}
    \includegraphics[width=\linewidth]{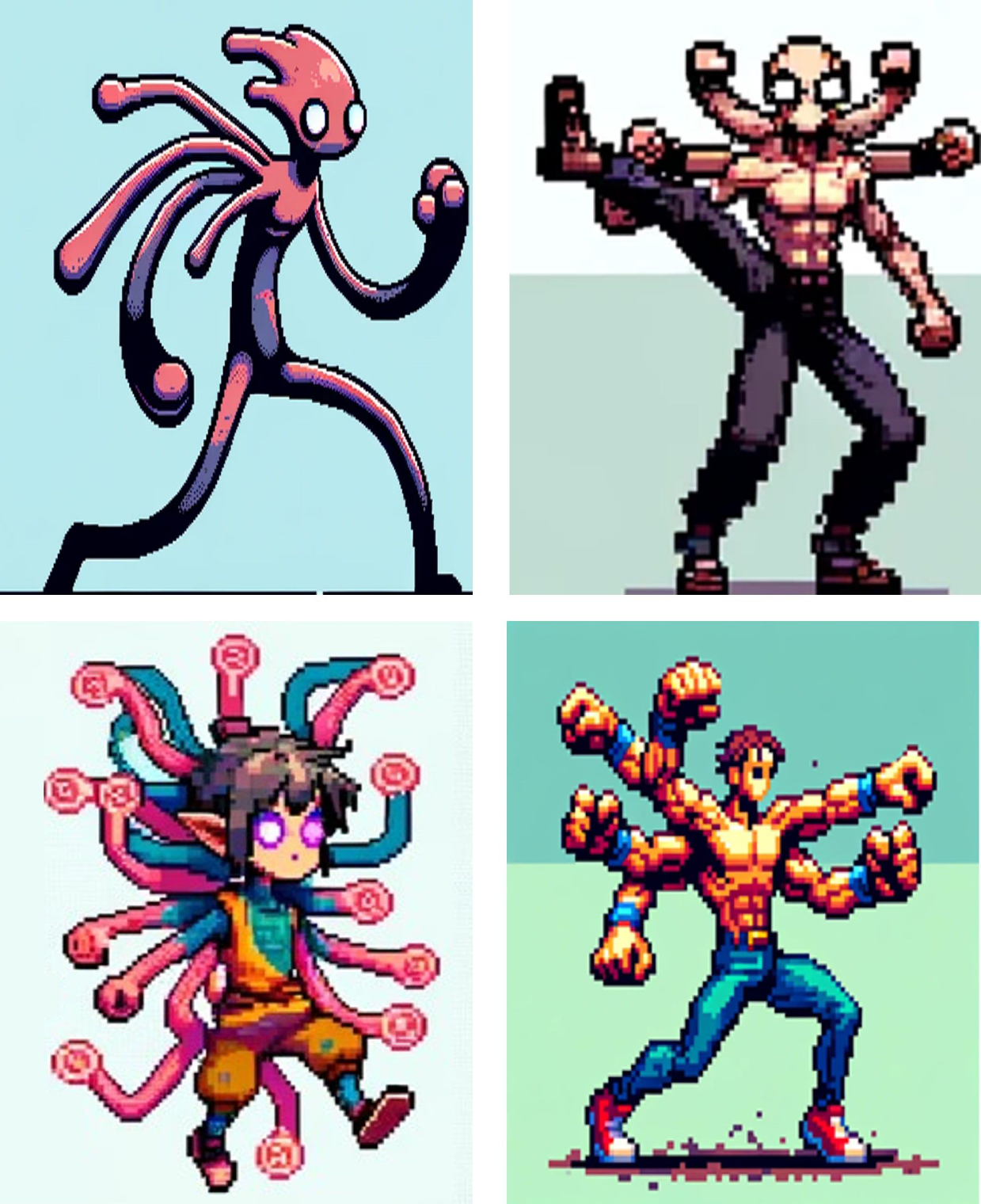}
    \caption{Fake hallucination.}
    \label{fig:fake_hallucination}
  \end{subfigure}
  \caption{Structural gap between real hallucination samples from normal prompt and fake hallucination samples from deliberately designed hallucination prompt. We highlight that generated fake hallucination samples (b) (intentionally generated by \textit{hallucination prompt}) is not enough to imitate real one, limiting to generate large data sample.}
  \label{fig:deliberate_example}
\end{figure}

Recent progress in the field of in-context visual learning has shown promising results in addressing various computer vision tasks. It is realized upon the advantages of in-context learning \cite{brown2020language} which is a prominent capability of model to specialize the existing VLMs or LLMs to desired new tasks rapidly within a context without separated training or tuning stage. The seminal work by Wang et al. \cite{wang2023images} introduced the concept of concatenating images to enable the application of a single model to multiple tasks. Zhang et al. \cite{zhang2024makes} further extended this approach by proposing a prompt retrieval framework to automatically select the most relevant examples for a given query, leading to improved performance. Additionally, integrating multiple example samples into a single grid image has been shown to enhance performance as the number of examples increases \cite{zhang2024makes}. Meanwhile, FAITHSCORE \cite{jing2023faithscore} proposed to evaluate various hallucination types with VLMs.

Nevertheless, these approaches have primarily focused on photorealistic images, and their effectiveness in non-photorealistic rendering (NPR) domains, such as cartoon characters, remains unexplored. Cartoon and pixel-style images, which represent non-photorealistic rendering \cite{wu2022make, kim2024toonaging, kim2022cross, back2022webtoonme, yang2022pastiche}, pose unique challenges due to their distinctive appearance and the difficulty in collecting large-scale datasets \cite{kim2024minecraft, yang2022pastiche, wu2022make}. Consequently, generating cartoon-style characters via TTI models often results in frequent \textit{semantic structural hallucination}s, such as characters with three legs or one arm, as shown in Fig. \ref{fig:hallucination_example}. To address this problem, a dedicated refinement or post-processing step is necessary. However, to the best of our knowledge, no research has been conducted to specifically tackle this issue in cartoon-style images.

To bridge this gap, we present a novel visual hallucination detection system based on few-shot pose-aware in-context visual learning (PA-ICVL) in NPR domains. Our approach takes inspiration from recent progress in the area of in-context visual learning \cite{wang2023images, zhang2024makes}. However, we extend these methods to address the unique challenges posed by cartoon character images by incorporating numerical pose information alongside the visual data, necessitating a novel adaptation of the existing techniques. In PA-ICVL, we do not perform additional parameter training or tuning on the given input-output pairs. Instead, we leverage a capability of in-context visual learning that gradually provides the model with both visual and pose information. This enables our model to effectively utilize the in-context learning paradigm for the specific new task of visual hallucination detection in non-photorealistic domains, allowing VLMs to make more accurate decisions in identifying visual hallucinations. Our contributions include:

\begin{enumerate}
    \item To the best of our knowledge, we first proposed visual hallucination detection system on non-photorealistic rendering domain, especially for machine-generated character images by text-to-image model, collecting a new public cartoon-hallucination dataset.
    \item Within a capability of in-context visual learning on VLMs, we improve the detection accuracy with few-shot paired samples including RGB image, hallucination label and human-readable prompt.
    \item To further enhance the detection performance, posture information is leveraged which demonstrate significant accuracy improvement by fine-tuning the pose estimator with pixel-cartoon domain images.
\end{enumerate}

\begin{figure*}[ht]
    \centering
    \includegraphics[width=\linewidth]{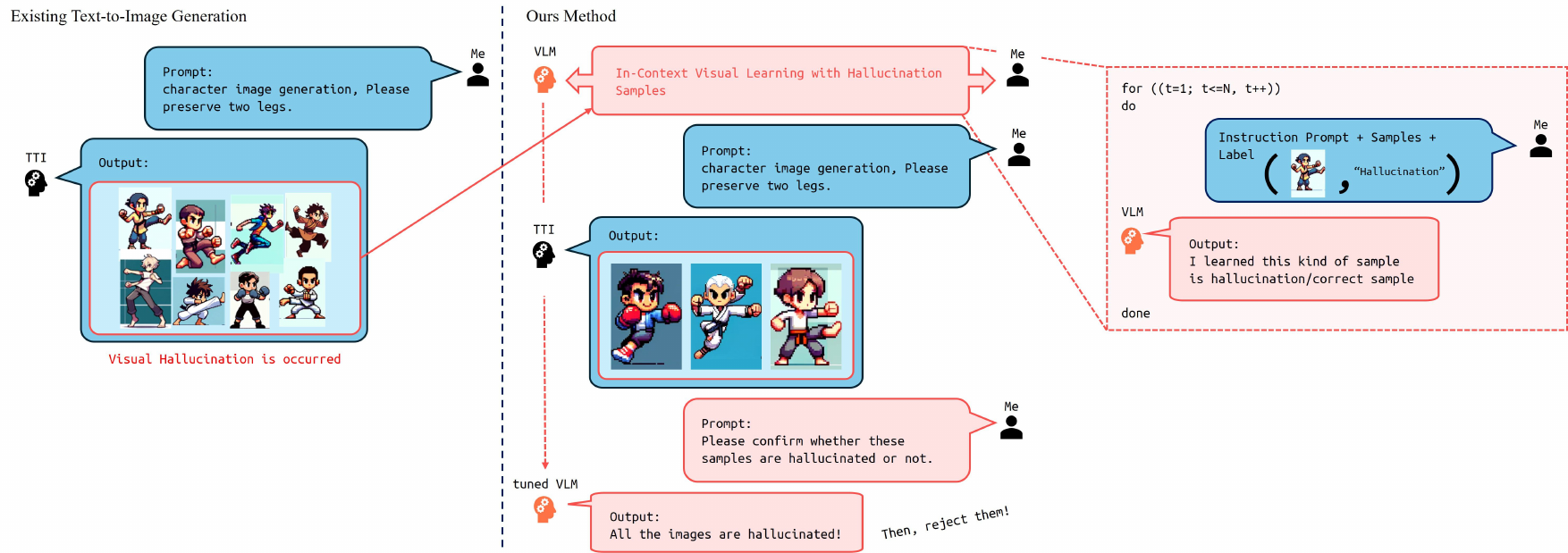}
    \hfil
\caption{Schematic comparison to leverage machine-generated images between using existing TTI process and using our proposed method with in-context learned VLM for hallucination verification.  {Our goal is to detect hallucinations in images generated from TTI using VLM.}}	
\label{fig:detection_scheme}    
\end{figure*}

\section{Challenges in Detecting Visual Hallucinations} \label{sec:challenges}

Before our proposal, we conducted pre-analysis and trials to detect visual hallucinations without the use of VLMs with in-context visual learning. Initially, we sought to employ image classification by generating a dataset with labels, yet we were confronted with a data imbalance issue due to the unpredictable and uneven generation of hallucination samples from the TTI process. In an effort to mitigate this challenge, we experimented with the deliberate generation of hallucination samples. However, this approach led us to face a gap in appearance between real and fabricated hallucinations. Through these endeavors, we concluded that amassing a large dataset for hallucination samples is not only limited but also a non-trivial task. This realization prompted us to explore the potential of learning from a few samples for the detection of hallucinations. In later sections, we provide some insights and detailed limitations through pre-analysis and trials to detect hallucination without VLMs.

\subsection{Class Imbalance Problem} \label{sec:class_imbalance}
Our first approach was to adopt image classification to detect it by generating hallucination dataset with label. To do that, we wanted to gather correct (\textit{i.e.}, non-hallucination) and hallucination samples through multiple TTI process. However we encountered data imbalance problem due to the fact that hallucination sample is unpredictably and unevenly generated with few samples from all the output according to textual prompts. Empirically, we faced about 10 hallucination samples per 100 images, but it is  {highly unpredictable} according to input prompt and settings. From these observations, we concluded that gathering hallucination samples with iterative TTI is burdensome and inefficient.

\subsection{Imitating Hallucination}
To address aforementioned class imbalance issue, we wanted to generate synthetic dataset for hallucination samples deliberately such as \cite{muqeet2023video, bai2023ffhq}  {with \textit{hallucination prompt}}, but faced appearance gap problem between real and fake hallucination as depicted in Fig. \ref{fig:deliberate_example}. We found that deliberately generated samples through our hallucination prompt has exaggerated structure (Fig. \ref{fig:fake_hallucination}.) much far ways than real hallucination (Fig. \ref{fig:real_hallucination}). Thus, we concluded that deliberately synthesizing large dataset for hallucination samples is also non-trivial and limited, and it makes us use few-samples for hallucination detection.

\section{Methodology}

Our goal is to recognize \textit{semantic structural hallucination} in generated image by TTI using a capability of in-context learning on VLMs, as illustrated in Fig. \ref{fig:detection_scheme}.  {To do that, we first collect hallucination-aware cartoon character dataset (Sec. {\ref{sec:visual_learning}}). Then, we demonstrate how posture information can be used in PA-ICVL (Sec. {\ref{subsec:pose_guidance}}). Finally, it is explained how VLM make the decision in hallucination detection step (sec. {\ref{subsec:hallucination_detection}})}

\subsection{Hallucination-aware Cartoon Dataset} \label{sec:visual_learning}

 {To conduct PA-ICVL within a VLM, we delicately collect the cartoon hallucination dataset. Concretely,} given input prompt $\textbf{P}_{\texttt{input}}$\footnote{We provide used prompt in Appendices \ref{suppl:input_prompt}.}, cartoon character image $\textbf{X}_\texttt{unknown}$ is generated as $\textbf{X}_\texttt{unknown} = f(\textbf{P}_{\texttt{input}})$ where $f(\cdot)$ is a TTI model. Through manual human annotation,  {label $\textbf{T}_*$ and description $\textbf{P}_{\texttt{desc}}$ are allocated while revealing $\textbf{X}_{\texttt{unknown}}$ to $\textbf{X}_{\texttt{known}}$}.  {By repeating this step, hallucination-aware} dataset $\mathcal{D}^{\text{H}}$ is validated as $\mathcal{D}^{\text{H}} = ([\textbf{X}^{\texttt{1}}_\texttt{known}, \textbf{T}^{\texttt{1}}_\texttt{*}, \textbf{P}^{\texttt{1}}_\texttt{desc}], \dots, [\textbf{X}^{\texttt{N}}_\texttt{known}, \textbf{T}^{\texttt{N}}_\texttt{*}, \textbf{P}^{\texttt{N}}_\texttt{desc}])$ where \texttt{N} is the number of samples. In each sample, $\textbf{P}^{\texttt{t}}_{\texttt{desc}}$ is $\texttt{t}$-th description prompt which includes  {the explanation about} why this sample is hallucinated/correct, $\textbf{T}_\texttt{*}^\texttt{t}$ is the $\texttt{t}$-th label where $*$ is \texttt{known} or \texttt{unknown}, and is obtained as $\textbf{T}^\texttt{t}_\texttt{h}=$ \texttt{"This is hallucinated one"} or $\textbf{T}^\texttt{t}_\texttt{c}=$ \texttt{"This is correct one"} for a hallucination and a correct sample, respectively.

\begin{algorithm}[t]
    \caption{\textsc{PA-ICVL}}
    \label{alg:tuning}
    \textbf{Input}: Hallucination dataset $\mathcal{D}^{\text{H}} =([\textbf{X}^{\texttt{1}}_{\texttt{known}},\textbf{M}^{\texttt{1}}_{\texttt{known}},\textbf{T}^{\texttt{1}}_{\texttt{*}}, \\ \textbf{P}^{\texttt{1}}_{\texttt{desc}}], \dots, [\textbf{X}^{\texttt{N}}_{\texttt{known}}, \textbf{M}^{\texttt{N}}_{\texttt{known}}, \textbf{T}^{\texttt{N}}_{\texttt{*}}, \textbf{P}^{\texttt{N}}_{\texttt{desc}}])$, system prompt $\textbf{P}_{\texttt{sys}}$, a general-purpose VLM $g(\cdot)$.\\
    \textbf{Output}: learned VLM $\hat{g}(\cdot)$
    \begin{algorithmic}[1]
        \State $\texttt{t} \gets 1$
        \State // \textit{Initialize VLM}
        \State  {$\hat{g}_{\texttt{0}}(\cdot) \gets g(\textbf{P}_{\texttt{sys}})$}
        \While{$\texttt{t} \neq \texttt{N+1}$}
            \State  {$\textbf{X}^{\texttt{t}}_{\texttt{known}}, \textbf{M}^{\texttt{t}}_{\texttt{known}}, \textbf{T}^{\texttt{t}}_{\texttt{*}}, \textbf{P}^{\texttt{t}}_{\texttt{desc}} \gets \mathcal{D}^{\text{H}}_{\texttt{t}}$}
            \State // \textit{PA-ICVL}
            \State  {$\hat{\textbf{T}}^{\texttt{t}}_{\texttt{*}} \gets \hat{g}_{\texttt{t-1}}(\textbf{X}^{\texttt{t}}_{\texttt{known}}, \textbf{M}^{\texttt{t}}_{\texttt{known}}, \textbf{P}^{\texttt{t}}_{\texttt{desc}})$}
            \If{$\hat{\textbf{T}}^{\texttt{t}}_{\texttt{*}} == \textbf{T}^{\texttt{t}}_{\texttt{*}}$}
                \State $\hat{g}_{\texttt{t}}(\cdot) \gets \hat{g}_{\texttt{t-1}}(\cdot)$
                \State $\texttt{t} \gets \texttt{t+1}$                
            \EndIf
        \EndWhile
        \State \textbf{return} $\hat{g}_{\texttt{N}}(\cdot)$
    \end{algorithmic}
\end{algorithm}

\subsection{Pose-Aware In-Context Visual Learning} \label{subsec:pose_guidance}

In the hallucination sample (Fig. \ref{fig:hallucination_example}), we found that visual hallucination samples from the cartoon domain include body structure issues such as arms, legs or heads. Based on these observations, we decide to leverage pose information to further improve the detection performance of VLM. 

Given pre-trained pose estimator $\mathcal{E}_\theta(\cdot)$ which has  {weights $\theta$ obtained by training from the cartoon-domain}, pose map $\textbf{M}$ is extracted as $\textbf{M}_* = \mathcal{E}_\theta(\textbf{X}_*)$  {where $\textbf{M}_*$ has the same resolution as the input image $\textbf{X}_*$ and contains the pose joint information as the channel-axis.} Thus, final each input of PA-ICVL for VLM is set as $[\textbf{X}^{\texttt{t}}_\texttt{known}, \textbf{M}^\texttt{t}_{\texttt{known}}, \textbf{T}^{\texttt{t}}_\texttt{*}, \textbf{P}^{\texttt{t}}_\texttt{desc}]$ where $\textbf{M}^\texttt{t}_{\texttt{known}}$ is \texttt{t}-th pose map from $\textbf{X}^{\texttt{t}}_\texttt{known}$. Each pose map $\textbf{M}$ contains pose information about the human structure,  {and the data structure changes depending on the experiments described in Sec. {\ref{subsec:ablation}}}. Details about pose estimator $\mathcal{E}_\theta(\cdot)$ is described in Appendices \ref{suppl:pose_estimator}. After PA-ICVL outlined in the Alg. \ref{alg:tuning},  {a hallucination-aware VLM $\hat{g}_{\texttt{N}}(\cdot)$ is obtained which is ready to detect visual hallucinations in inference step (Sec. {\ref{subsec:hallucination_detection}}).}

\begin{figure}[t]
\centering
    \includegraphics[width=1.0\linewidth]{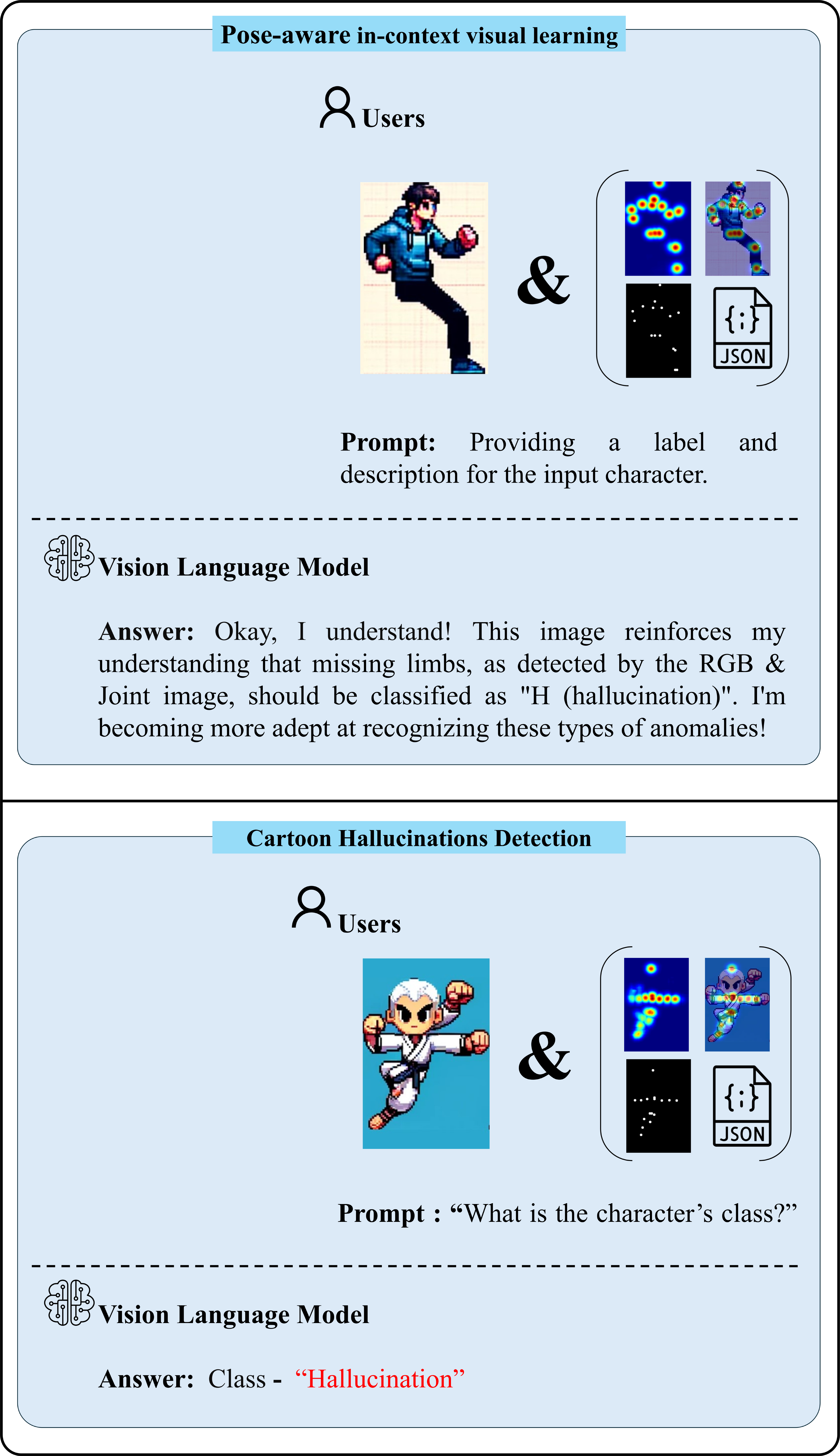}
\caption{Example of PA-ICVL step  {(top)} and detection step  {(bottom)}.}
\vspace{-1.5em}
\label{fig:example_pa_icvl_process}   
\end{figure}

\begin{figure*}[t]
\centering
    \includegraphics[width=\linewidth]{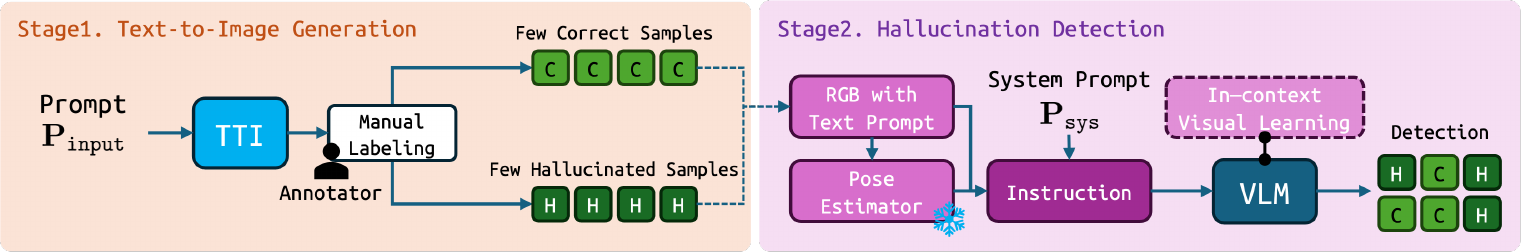}
    \hfil
\caption{Pipeline of  {cartoon-hallucination dataset collection (Stage 1)} and hallucination detection  {(Stage 2)}.}	
\label{fig:overall_pipeline}
\end{figure*}

\subsection{Visual Hallucination Detection} \label{subsec:hallucination_detection}

 {Till now, we explained how VLM learn visual hallucination based on in-context learning without updating parameters. Here, we describe how the hallucination-aware VLM can be utilized when faced with hallucination-unknown samples. Let's say} newly generated image $\textbf{X}_\texttt{unknown}$  {is generated by user using TTI model}, the pose estimator first $\mathcal{E}_\theta(\cdot)$ extracts a pose map as $\textbf{M}_\texttt{unknown} = \mathcal{E}_\theta(\textbf{X}_\texttt{unknown})$. Then,  {hallucination-aware VLM} $\hat{g}_{\texttt{N}}(\cdot)$ predict hallucination label $\hat{\textbf{T}}_\texttt{p}$ as $\hat{\textbf{T}}_\texttt{p} = \hat{g}_{\texttt{N}}(\textbf{X}_{\texttt{unknown}}, \textbf{M}_\texttt{unknown})$. Finally,  {hallucination-unknown} one $\textbf{X}_\texttt{unknown}$ is determined as :

\begin{equation}
\textbf{X}_\texttt{unknown} =\begin{cases}
 \hat{\textbf{X}}_{\texttt{known,c}} & \text{ if } \hat{\textbf{T}}_\texttt{p} = \textbf{T}_\texttt{c}\\ 
 \hat{\textbf{X}}_{\texttt{known,h}} & \text{ if } \hat{\textbf{T}}_\texttt{p} = \textbf{T}_\texttt{h}
\end{cases}
\end{equation}

 {Example of PA-ICVL with detection is shown in Fig. {\ref{fig:example_pa_icvl_process}}.} After detecting all the samples in new hallucination-unknown dataset, this dataset $\{\textbf{X}_{\texttt{unknown}}\}$ is validated as hallucination-known dataset $\{\hat{\textbf{X}}_{\texttt{known,c}}\}$, \textit{i.e.}, non-hallucinated samples with different the number of samples. Overall pipeline for hallucination detection process is illustrated in Fig. \ref{fig:overall_pipeline}.

\section{Experiments}
\subsection{Experimental Settings}
For character appearance, we limit the style as cartoon and pixel using prompt  {which include an expression about human-like structure with five-head-figure}\footnote{ {Note that descriptions of non-human-like cartoon characters are in the Appendices {\ref{supple:non_human_like}}.}}. For TTI model, we adopt DALL-E3\footnote{Note that we found meaningful tendency between ChatGPT platform and pure DALL-E3 API. We invite reader to Appendices \ref{supple:dalle_api} for details.} \cite{betker2023improving}. This is because we found that other TTI models (SDv1.5 \cite{rombach2022high}, PixArt-$\alpha$ \cite{chen2023pixartalpha, chen2024pixartdelta} and so on) cannot generate consistent appearance (see Appendices \ref{suppl:t2i_comaprison} for TTI comparison). Final TTI image $\textbf{X}$ and input size of pose estimation has 384 by 256. For pose information, we used MPII \cite{andriluka14cvpr} format for joint labeling which is represented as 16 joint feature as raw pose map $\text{M}$. Details on fine-tuning pose estimator are included in Appendices \ref{suppl:fine_tuning}. For VLMs, GPT-4.0 Vision \cite{achiam2023gpt}, Gemini 1.5 Pro \cite{reid2024gemini} are selected.

\subsection{Quantitative Evaluation with Ablation} \label{subsec:ablation}

\begin{figure*}[t]
\centering
    \includegraphics[width=0.80\linewidth]{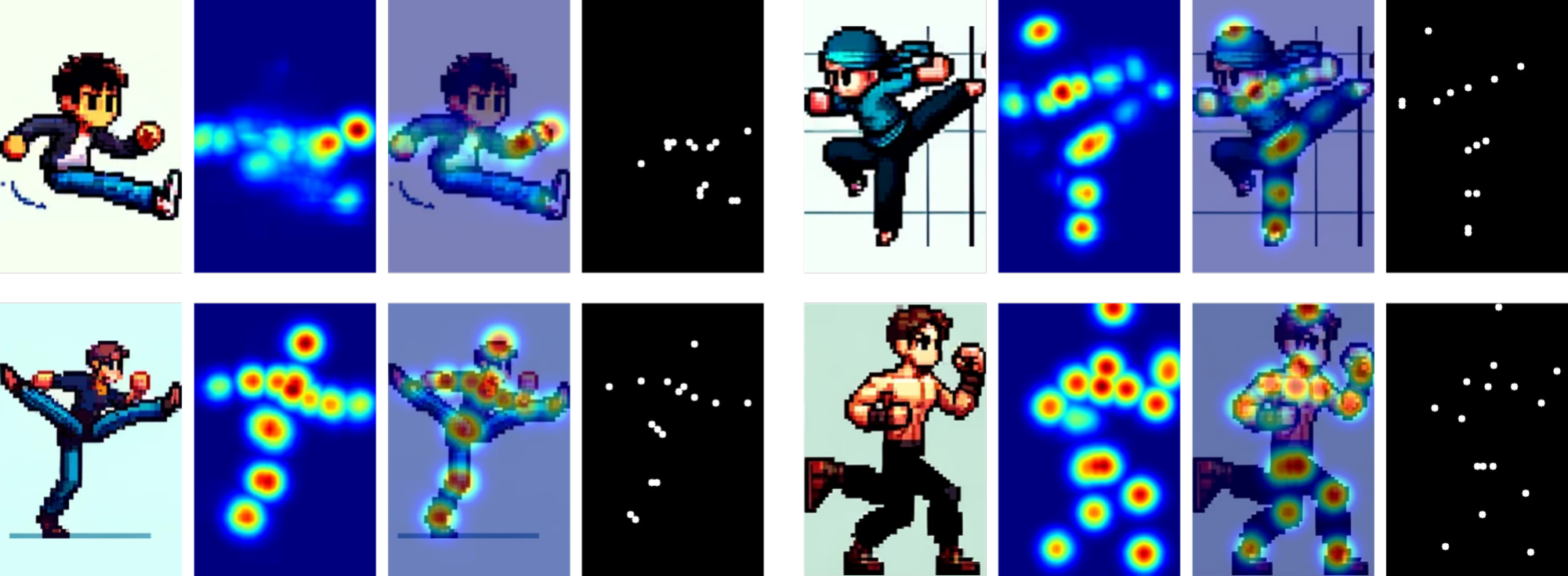}
\caption{Samples of our cartoon-hallucination dataset.}
\label{fig:dataset_samples}
\end{figure*}

VLM learned our state using the 5 correct and 5 hallucination train samples (total $\texttt{N}=10$ images for PA-ICVL)\footnote{ {Note that the results for the number of $\texttt{N}$ are in the Appendices {\ref{supple:fewer_samples}},}} and detect the test samples using 60 images for each class (total 120 images for evaluation) in our dataset. Dataset samples are shown in Fig. \ref{fig:dataset_samples}. To quantitatively evaluate the performance, we calculated the number of correct predictions relative to the total number of test samples for each class. This metric provides a clear indication of the learned VLM's ability to accurately detect visual hallucinations in machine-generated pixel character images.

\begin{figure}[t]
  \centering
  \begin{subfigure}[t]{0.22\linewidth}
    \includegraphics[width=\linewidth]{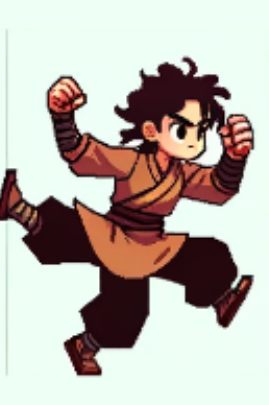}
    \caption{RGB $\textbf{X}$}
    \label{fig:rgb}
  \end{subfigure}
  \begin{subfigure}[t]{0.22\linewidth}
    \includegraphics[width=\linewidth]{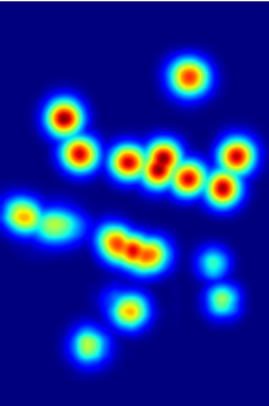}
    \caption{Gaussian heatmap $\textbf{M}_{gau}$}
    \label{fig:heatmap}
  \end{subfigure}
  \begin{subfigure}[t]{0.22\linewidth}
    \includegraphics[width=\linewidth]{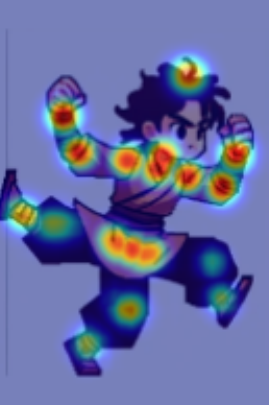}
    \caption{Overlapped gaussian heatmap $\textbf{M}_{over}$}
    \label{fig:overlap_heatmap}
  \end{subfigure}
  \begin{subfigure}[t]{0.22\linewidth}
    \includegraphics[width=\linewidth]{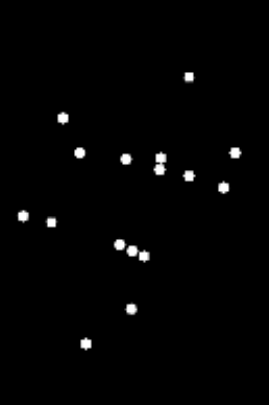}
    \caption{Final joint $\textbf{M}_{xy}$}
    \label{fig:finaljoint}
  \end{subfigure}
  \caption{Example of input data about VLM for ablation study.}
  \label{fig:pose_vis}
\end{figure}

To verify the effectiveness of our method, we conduct ablation study as shown in Tab. \ref{tab:ablation}.  Model \textbf{A} is system prompt only VLMs. Model \textbf{B} is system prompt with the definition about hallucination. Model \textbf{C} is visual in-context learning. Multiple models \textbf{D} are our methods, PA-ICVL with various pose map format as shown in Fig. \ref{fig:pose_vis}.  {$\textbf{M}_{gau}$ is Gaussian heatmap extracted from final convolution layer of pose estimator, $\textbf{M}_{over}$ is an overlay image using $\textbf{X}$ and $\textbf{M}_{gau}$, $\textbf{M}_{xy}$ is final output of the pose estimator, which has the maximum value of each channel of $\textbf{M}_{gau}$ as the $(x, y)$ coordinates.}

\begin{table}[h]
    \centering
    \begin{tabular}{l|l|l} 
         \toprule
         Model & Task Name & Inputs for VLM  \\
         \midrule
         \textbf{A} & System Prompt & . \\ 
         \textbf{B} & \textbf{A} + Hallucination Define. & . \\ 
         \textbf{C} & \textbf{B} + In-Context Learning & $\textbf{X}$ \\ 
         \midrule
         $\textbf{D}$ & \multicolumn{1}{c|}{Pose Guidance} &  \\ 
         $\textbf{D-1}  $ & \textbf{C} + Gaussian Heatmap & $ \textbf{X}, \textbf{M}_{gau}$  \\ 
         $\textbf{D-2}  $ & \textbf{C} + Only Overlapped Heatmap & $ \textbf{M}_{over}$  \\ 
         $\textbf{D-3}  $ & \textbf{C} +  Overlapped Heatmap & $ \textbf{X}, \textbf{M}_{over}$  \\ 
         $\textbf{D-4}  $ & \textbf{C} + Joint (image) & $ \textbf{X}, \textbf{M}_{xy}$  \\ 
         $\textbf{D-5}^\dagger $ & \textbf{C} + Joint (text) & $ \textbf{X}, \texttt{text}(\textbf{M}_{xy})$  \\
         \bottomrule
         
    \end{tabular}
    \caption{Ablation list. $\dagger$ denotes our final model}
    \label{tab:ablation}
\end{table}

\begin{figure}[t]
\centering
    \includegraphics[width=1.0\linewidth]{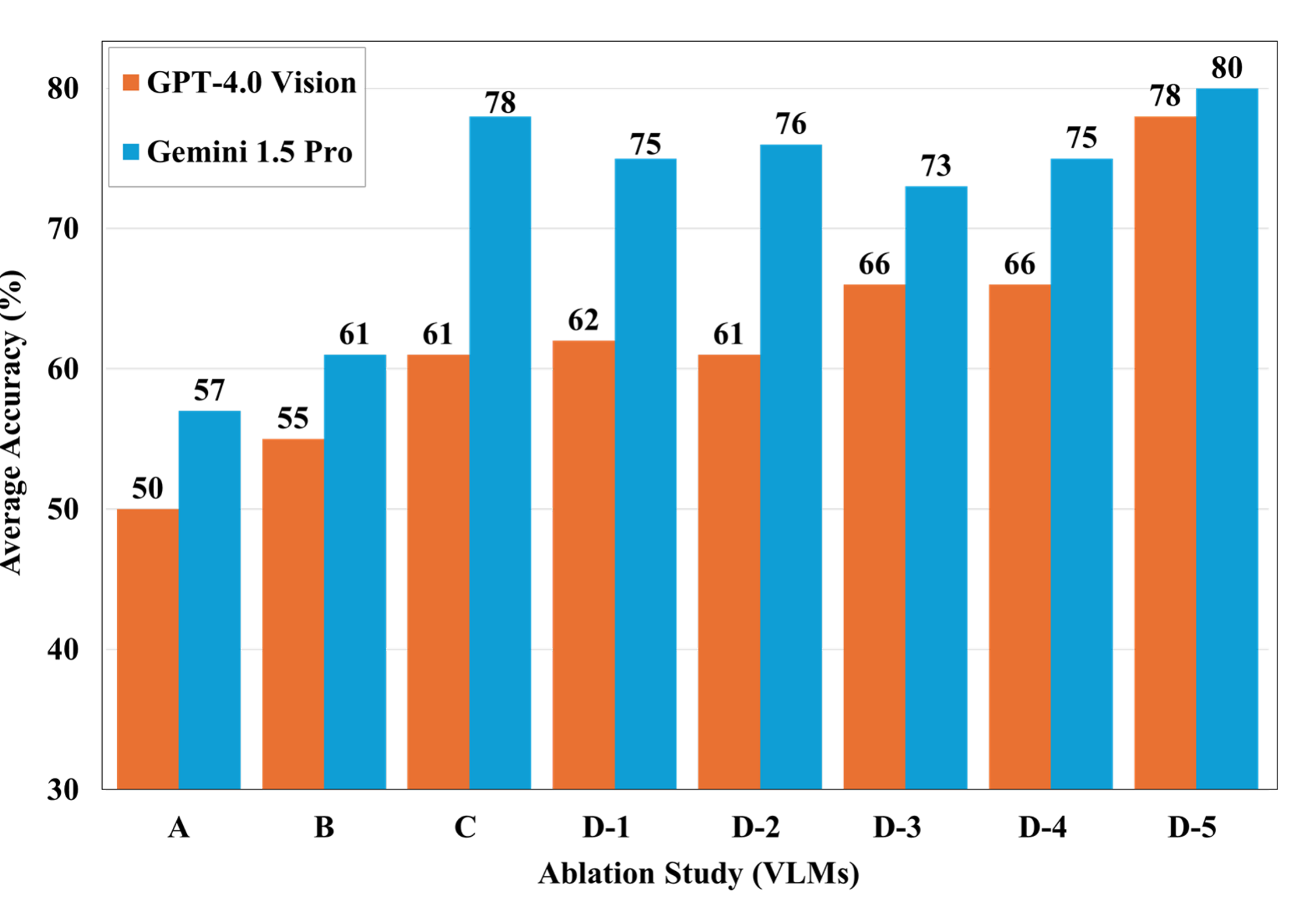}
\caption{Quantitative evaluation with ablation study about \textit{semantic structural hallucination} detection.}
\label{fig:quantitative_results}    
\end{figure}

From ablation results in Fig. \ref{fig:quantitative_results}, our quantitative results demonstrate that model \textbf{A} seems to detect inputs randomly with about 50\% accuracy, while model \textbf{B} shows a better score than model \textbf{A}, but still performs poorly. In contrast, model \textbf{C}, which learned from our hallucination cases, exhibits significantly better detection performance in both VLMs. Building upon this foundation, pose-aware models \textbf{D-1}, \textbf{D-2}, \textbf{D-3} and \textbf{D-4} demonstrate improved performance thanks to the additional pose input within GPT-4.0 Vision. In contrast, these additional inputs $\textbf{M}_{gau}, \textbf{M}_{over}$ and $\textbf{M}_{xy}$ seems like to disturb the interpretation process in the case of Gemini 1.5 Pro with rather lower score. We conjecture this tendency is derived from that visual encoder equipped in GPT-4.0 Vision has a more prominent capability to understand various RGB image excluding general appearance than those of Gemini 1.5 Pro. Beyond using the image modality for pose, the language modality-based pose-aware model \textbf{D-5} achieves the best score in both VLMs. We argue that textual joint data can provide more precise posture information, enabling the VLM to effectively compare the input RGB image with the pose data to identify hallucinations.

\begin{table}[t]
    \centering
    \begin{tabular}{c|c|c|c}
        \toprule
        Method & Subject  & Approach  & Cost per infer \\
        \midrule
        Ours   & Computer & PA-ICVL   & 663 tokens \& 3 sec             \\
        Manual & Human    & heuristic & None \& 45 sec  \\
        \bottomrule
    \end{tabular}
    \caption{Cost approximation in detecting hallucination. Note that the cost of the proposed method is based on the average number of tokens (input tokens + output tokens) and the time it takes to inference, while the cost of the manual approach include $\textbf{P}_{\texttt{desc}}$.}
    \label{tab:cost}
\end{table}

\subsection{Cost Prediction of Hallucination Detection}

 {In terms of cost-effective feasibility, we compared the estimated detection costs of our proposed method against manual visual labeling. There is no existing method to detect cartoon hallucination, and training hallucination models with large samples is challenging, as discussed in Sec. {\ref{sec:challenges}}. Therefore, we only compared our method to manual labeling based on human perception.}

 {In our PA-ICVL method, the average input to infer one character consists of about 513 tokens, assuming an RGB image and keypoint JSON file combined with a prompt (e.g., `What is this character’s class?'). The images used in our experiments were 384x256 pixels, with each image using 255 tokens and the corresponding keypoint JSON file using 258 tokens. The total input tokens for inferring 120 characters amounted to 62.7K. VLM outputs an average of 140 tokens per character for classification and description, totaling about 17K tokens for all characters. Inference for the entire dataset took approximately 10 minutes, excluding the 1-2 hour fine-tuning step of the pose estimator on an NVIDIA RTX 3090. After training, inference for our whole dataset using the pose estimation model took just 25 seconds.}

 {For the manual detection process, two participants classified the same characters. They were asked to identify the images, classify them into separate folders, and add captions $\textbf{P}_{\texttt{desc}}$ explaining why the sample was hallucinated/corrected. Each inference step for this took an average of 45 seconds, and the entire process lasted approximately 1 hour and 30 minutes. No hallucination samples were provided, and participants were only given a definition of hallucination based on Model \textbf{B}. Overall comparison between two approaches is shown in Tab. {\ref{tab:cost}}.}

\section{Discussion}

This section, we discuss about applications and extendability of our proposed method, hoping that it can serve as some insights or inspirations for multi-modal communities and other researchers.

Our domain, cartoon style character, has extremely wide appearance. What is more, some cases include ambiguous appearance even in human cognition. Therefore, our learned VLM can not detect all the images of wide-range cartoon style. Meanwhile, in the perspective of model capabilities, it is known that TTI models and VLMs tend to generate output sensitively with word position-bias, verbosity-bias, self-enhancement-bias \cite{zheng2024judging}. Due to that, it is noteworthy that output or results can be showed as quite different tendency according to the inputs. Through above mentioned things, we would like to mention that there are some limitations to leverage our scheme toward more generalized task directly. Following description include our attempts and faced limitations with some specific application cases.

\begin{figure}[t]
\centering
    \includegraphics[width=1.0\linewidth]{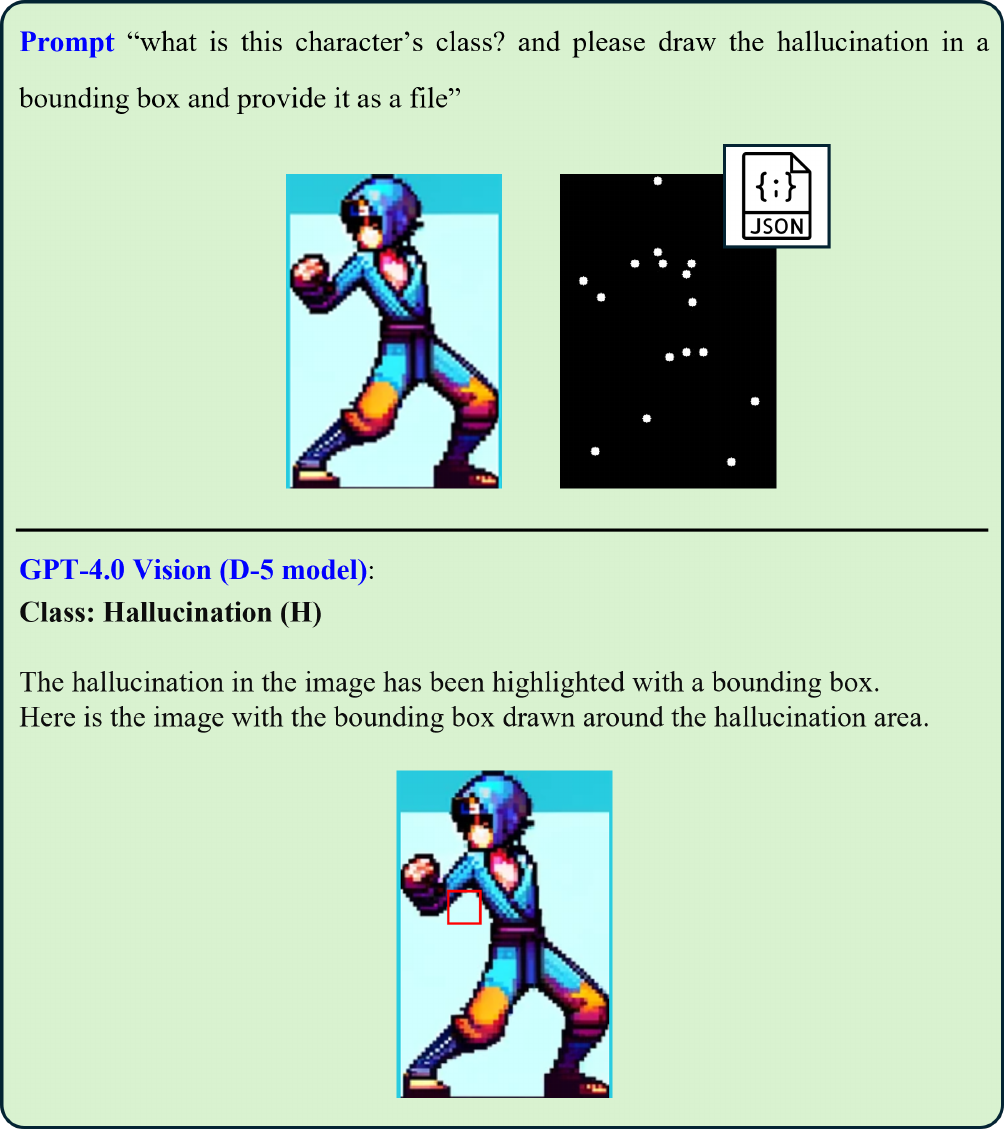}
\caption{Visualization of hallucination region localization of VLM. \textcolor{red}{Red} bounding box denotes what VLM extract as dominant region to detect \textit{semantic structural hallucination}.}
\vspace{-1.0em}
\label{fig:localization}    
\end{figure}

\begin{figure}[t]
\centering
    \includegraphics[width=1.0\linewidth]{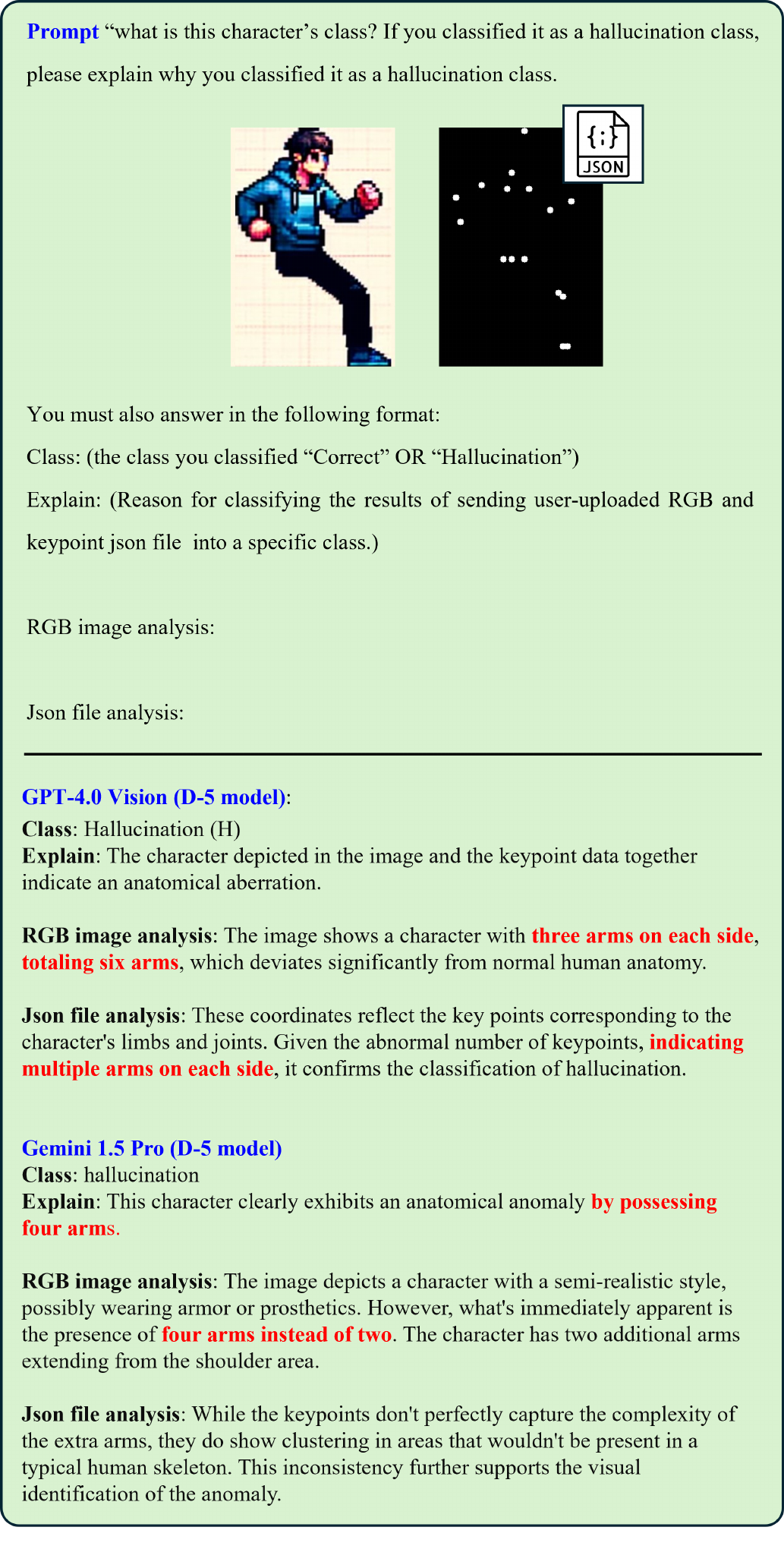}
\caption{Explanations about the reason of hallucination. We denotes wrong explanation as \textbf{\textcolor{red}{Red-bold}} font.}
\label{fig:Limit2}    
\end{figure}

\subsection{Limitation: Hallucination-Region Localization} \label{sec:limit1}
We wanted to exploit hallucination region toward several applications such as regional restoration by feeding finding region as desired masking map (e.g., in-painting \cite{lugmayr2022repaint}). To do this, we let the tuned VLM extract and visualize the structural hallucination region as rectangular bounding box.

In toy experiments as shown in Fig. \ref{fig:localization}, we observed that VLM failed to extract hallucination elements in given images. To improve this, it can be considered that in-context visual learning is conducted with an annotated bounding box information of hallucination elements. We note that Gemini 1.5 Pro is excluded in this test since they have not yet that function for drawing bounding box.

\subsection{Limitation: Explainability}

As a continuation of Sec. \ref{sec:limit1}, we also would like to explore the explanability of VLMs about hallucination detection. By doing that, we expect to understand how VLMs reason the hallucination within given samples beyond the classification. Here, we instructed the VLMs to explain the detailed reason about their decision as a textual prompt. As shown in Fig. \ref{fig:Limit2}, it is observed that description capability of each VLM is insufficient to explain the hallucination elements. Therefore, as a result, we mention that explaining the reason via VLMs can not always ensure the correct answers.

\subsection{Future Works}

In this paper, the range of visual hallucination appearance was limited as character structure with five-head-figure. However, more natural scene (\textit{e.g.}, video or 3D) for real-world contain wide structure. For instance, some images has not contain full-body. For future work, we intent to detect partial and detailed hallucination like fingers and would like to realize restoration about hallucination region based on region localization as discussed in Sec. \ref{sec:limit1}. Moreover, generalizability for realistic images will be included in advanced version of our works.

\section{Conclusion}

Our research has introduced a novel visual hallucination detection system for cartoon character images generated by large-scale TTI models. To address this issue, inspired by recent progress in in-context visual learning, we leveraged PA-ICVL, integrating RGB images with pose information from a fine-tuned pose estimator. This method extends existing in-context visual learning strategies by adding numerical pose data to visual inputs, utilizing a distinctive repetitive information injection approach. By iteratively feeding visual and pose data into the model within a contextual history, our system leverages the advantages of in-context learning with no additional training. Moreover, we collected a cartoon-hallucination dataset with corresponding pose maps which will be publicly available. As a result, VLMs have become more precise in identifying visual hallucinations. Our experiments have demonstrated that our approach significantly improves upon baseline methods, notably enhancing TTI models by mitigating visual hallucinations and broadening their applicability, particularly in scenarios where visual accuracy is paramount. Specifically, through testing under various pose information conditions, we found that incorporating language modal-based joint information proved to be the most effective strategy. In addition, our experiments showed that external condition can reinforce the visual understanding capabilities of VLM by improving task-specific performance, achieving 78\%, 80\% detection performance from 50\%, 57\% about GPT-4v, Gemini pro vision, respectively. These experimental results convey the significant insight where general VLM can be specialized to each domain by taking advantages of in-context visual learning with external information. Future work will aim to integrate additional modalities and adapt our approach to various non/photorealistic styles, enhancing TTI model applications. Our method, addressing cartoon image challenges, establishes new benchmarks in detecting visual hallucinations in such domains.

\section*{Acknowledgments}

This research was supported by Culture, Sports and Tourism R\&D Program through the Korea Creative Content Agency grant funded by Ministry of Culture, Sports and  Tourism in 2024 (Project Name : Developing Professionals for R\&D in Contents Production Based on Generative Ai and Cloud, Project Number : RS-2024-00352578, Contribution Rate: 50\%) and Culture, Sports and Tourism R\&D Program through the Korea Creative Content Agency (KOCCA) grant funded by the Ministry of Culture, Sports and Tourism (MCST) in 2023 (Project Name: Development of digital abusing detection and management technology for a safe Metaverse service, Project Number: RS-2023-00227686, Contribution Rate: 50\%) and Artificial intelligence industrial convergence cluster development project funded by the Ministry of Science and ICT(MSIT, Korea) \& Gwangju Metropolitan City.

{\small
\bibliographystyle{ieee_fullname}
\bibliography{egbib}

@article{rahmanzadehgervi2024vision,
  title={Vision language models are blind},
  author={Rahmanzadehgervi, Pooyan and Bolton, Logan and Taesiri, Mohammad Reza and Nguyen, Anh Totti},
  journal={arXiv preprint arXiv:2407.06581},
  year={2024}
}

@inproceedings{rombach2022high,
  title={High-resolution image synthesis with latent diffusion models},
  author={Rombach, Robin and Blattmann, Andreas and Lorenz, Dominik and Esser, Patrick and Ommer, Bj{\"o}rn},
  booktitle={Proceedings of the IEEE/CVF conference on computer vision and pattern recognition},
  pages={10684--10695},
  year={2022}
}

@article{nichol2021glide,
  title={Glide: Towards photorealistic image generation and editing with text-guided diffusion models},
  author={Nichol, Alex and Dhariwal, Prafulla and Ramesh, Aditya and Shyam, Pranav and Mishkin, Pamela and McGrew, Bob and Sutskever, Ilya and Chen, Mark},
  journal={arXiv preprint arXiv:2112.10741},
  year={2021}
}

@inproceedings{ramesh2021zero,
  title={Zero-shot text-to-image generation},
  author={Ramesh, Aditya and Pavlov, Mikhail and Goh, Gabriel and Gray, Scott and Voss, Chelsea and Radford, Alec and Chen, Mark and Sutskever, Ilya},
  booktitle={International Conference on Machine Learning},
  pages={8821--8831},
  year={2021},
  organization={PMLR}
}

@article{podell2023sdxl,
  title={Sdxl: Improving latent diffusion models for high-resolution image synthesis},
  author={Podell, Dustin and English, Zion and Lacey, Kyle and Blattmann, Andreas and Dockhorn, Tim and M{\"u}ller, Jonas and Penna, Joe and Rombach, Robin},
  journal={arXiv preprint arXiv:2307.01952},
  year={2023}
}

@article{blattmann2023stable,
  title={Stable video diffusion: Scaling latent video diffusion models to large datasets},
  author={Blattmann, Andreas and Dockhorn, Tim and Kulal, Sumith and Mendelevitch, Daniel and Kilian, Maciej and Lorenz, Dominik and Levi, Yam and English, Zion and Voleti, Vikram and Letts, Adam and others},
  journal={arXiv preprint arXiv:2311.15127},
  year={2023}
}

@article{zhang2023styleavatar3d,
  title={StyleAvatar3D: Leveraging Image-Text Diffusion Models for High-Fidelity 3D Avatar Generation},
  author={Zhang, Chi and Chen, Yiwen and Fu, Yijun and Zhou, Zhenglin and Yu, Gang and Wang, Billzb and Fu, Bin and Chen, Tao and Lin, Guosheng and Shen, Chunhua},
  journal={arXiv preprint arXiv:2305.19012},
  year={2023}
}

@article{achiam2023gpt,
  title={Gpt-4 technical report},
  author={Achiam, Josh and Adler, Steven and Agarwal, Sandhini and Ahmad, Lama and Akkaya, Ilge and Aleman, Florencia Leoni and Almeida, Diogo and Altenschmidt, Janko and Altman, Sam and Anadkat, Shyamal and others},
  journal={arXiv preprint arXiv:2303.08774},
  year={2023}
}

@article{liu2024visual,
  title={Visual instruction tuning},
  author={Liu, Haotian and Li, Chunyuan and Wu, Qingyang and Lee, Yong Jae},
  journal={Advances in neural information processing systems},
  volume={36},
  year={2024}
}

@article{alayrac2022flamingo,
  title={Flamingo: a visual language model for few-shot learning},
  author={Alayrac, Jean-Baptiste and Donahue, Jeff and Luc, Pauline and Miech, Antoine and Barr, Iain and Hasson, Yana and Lenc, Karel and Mensch, Arthur and Millican, Katherine and Reynolds, Malcolm and others},
  journal={Advances in Neural Information Processing Systems},
  volume={35},
  pages={23716--23736},
  year={2022}
}

@inproceedings{wang2023images,
  title={Images speak in images: A generalist painter for in-context visual learning},
  author={Wang, Xinlong and Wang, Wen and Cao, Yue and Shen, Chunhua and Huang, Tiejun},
  booktitle={Proceedings of the IEEE/CVF Conference on Computer Vision and Pattern Recognition},
  pages={6830--6839},
  year={2023}
}

@article{zhang2024makes,
  title={What makes good examples for visual in-context learning?},
  author={Zhang, Yuanhan and Zhou, Kaiyang and Liu, Ziwei},
  journal={Advances in Neural Information Processing Systems},
  volume={36},
  year={2024}
}

@inproceedings{yang2022pastiche,
  title={Pastiche master: Exemplar-based high-resolution portrait style transfer},
  author={Yang, Shuai and Jiang, Liming and Liu, Ziwei and Loy, Chen Change},
  booktitle={Proceedings of the IEEE/CVF Conference on Computer Vision and Pattern Recognition},
  pages={7693--7702},
  year={2022}
}

@inproceedings{sun2019deep,
  title={Deep high-resolution representation learning for human pose estimation},
  author={Sun, Ke and Xiao, Bin and Liu, Dong and Wang, Jingdong},
  booktitle={Proceedings of the IEEE/CVF conference on computer vision and pattern recognition},
  pages={5693--5703},
  year={2019}
}

@article{hirschorn2023pose,
  title={Pose Anything: A Graph-Based Approach for Category-Agnostic Pose Estimation},
  author={Hirschorn, Or and Avidan, Shai},
  journal={arXiv preprint arXiv:2311.17891},
  year={2023}
}

@article{zhang2023vision,
  title={Vision-language models for vision tasks: A survey},
  author={Zhang, Jingyi and Huang, Jiaxing and Jin, Sheng and Lu, Shijian},
  journal={arXiv preprint arXiv:2304.00685},
  year={2023}
}

@article{zhao2023survey,
  title={A survey of large language models},
  author={Zhao, Wayne Xin and Zhou, Kun and Li, Junyi and Tang, Tianyi and Wang, Xiaolei and Hou, Yupeng and Min, Yingqian and Zhang, Beichen and Zhang, Junjie and Dong, Zican and others},
  journal={arXiv preprint arXiv:2303.18223},
  year={2023}
}

@article{kim2024toonaging,
  title={ToonAging: Face Re-Aging upon Artistic Portrait Style Transfer},
  author={Kim, Bumsoo and Muqeet, Abdul and Lee, Kyuchul and Seo, Sanghyun},
  journal={arXiv preprint arXiv:2402.02733},
  year={2024}
}

@article{kim2024minecraft,
  title={Minecraft-ify: Minecraft Style Image Generation with Text-guided Image Editing for In-Game Application},
  author={Kim, Bumsoo and Byun, Sanghyun and Jung, Yonghoon and Shin, Wonseop and Amin, Sareer UI and Seo, Sanghyun},
  journal={arXiv preprint arXiv:2402.05448},
  year={2024}
}

@article{wu2022make,
  title={Make your own sprites: Aliasing-aware and cell-controllable pixelization},
  author={Wu, Zongwei and Chai, Liangyu and Zhao, Nanxuan and Deng, Bailin and Liu, Yongtuo and Wen, Qiang and Wang, Junle and He, Shengfeng},
  journal={ACM Transactions on Graphics (TOG)},
  volume={41},
  number={6},
  pages={1--16},
  year={2022},
  publisher={ACM New York, NY, USA}
}

@article{touvron2023llama,
  title={Llama: Open and efficient foundation language models},
  author={Touvron, Hugo and Lavril, Thibaut and Izacard, Gautier and Martinet, Xavier and Lachaux, Marie-Anne and Lacroix, Timoth{\'e}e and Rozi{\`e}re, Baptiste and Goyal, Naman and Hambro, Eric and Azhar, Faisal and others},
  journal={arXiv preprint arXiv:2302.13971},
  year={2023}
}

@article{huang2023survey,
  title={A survey on hallucination in large language models: Principles, taxonomy, challenges, and open questions},
  author={Huang, Lei and Yu, Weijiang and Ma, Weitao and Zhong, Weihong and Feng, Zhangyin and Wang, Haotian and Chen, Qianglong and Peng, Weihua and Feng, Xiaocheng and Qin, Bing and others},
  journal={arXiv preprint arXiv:2311.05232},
  year={2023}
}

@article{li2023evaluating,
  title={Evaluating object hallucination in large vision-language models},
  author={Li, Yifan and Du, Yifan and Zhou, Kun and Wang, Jinpeng and Zhao, Wayne Xin and Wen, Ji-Rong},
  journal={arXiv preprint arXiv:2305.10355},
  year={2023}
}

@inproceedings{hu2022scaling,
  title={Scaling up vision-language pre-training for image captioning},
  author={Hu, Xiaowei and Gan, Zhe and Wang, Jianfeng and Yang, Zhengyuan and Liu, Zicheng and Lu, Yumao and Wang, Lijuan},
  booktitle={Proceedings of the IEEE/CVF conference on computer vision and pattern recognition},
  pages={17980--17989},
  year={2022}
}

@article{kim2022cross,
  title={Cross-domain style mixing for face cartoonization},
  author={Kim, Seungkwon and Gwak, Chaeheon and Kim, Dohyun and Lee, Kwangho and Back, Jihye and Ahn, Namhyuk and Kim, Daesik},
  journal={arXiv preprint arXiv:2205.12450},
  year={2022}
}

@incollection{back2022webtoonme,
  title={Webtoonme: A data-centric approach for full-body portrait stylization},
  author={Back, Jihye and Kim, Seungkwon and Ahn, Namhyuk},
  booktitle={SIGGRAPH Asia 2022 Technical Communications},
  pages={1--4},
  year={2022},
  publisher={ACM}
}

@article{betker2023improving,
  title={Improving image generation with better captions},
  author={Betker, James and Goh, Gabriel and Jing, Li and Brooks, Tim and Wang, Jianfeng and Li, Linjie and Ouyang, Long and Zhuang, Juntang and Lee, Joyce and Guo, Yufei and others},
  journal={Computer Science. https://cdn. openai. com/papers/dall-e-3. pdf},
  volume={2},
  number={3},
  pages={8},
  year={2023}
}

@misc{kakaobrain2022karlo-v1-alpha,
  title         = {Karlo-v1.0.alpha on COYO-100M and CC15M},
  author        = {Donghoon Lee and Jiseob Kim and Jisu Choi and Jongmin Kim and Minwoo Byeon and Woonhyuk Baek and Saehoon Kim},
  year          = {2022},
  howpublished  = {\url{https://github.com/kakaobrain/karlo}},
}

@article{8765346,
  author = {Z. {Cao} and G. {Hidalgo Martinez} and T. {Simon} and S. {Wei} and Y. A. {Sheikh}},
  journal = {IEEE Transactions on Pattern Analysis and Machine Intelligence},
  title = {OpenPose: Realtime Multi-Person 2D Pose Estimation using Part Affinity Fields},
  year = {2019}
}

@article{zheng2024judging,
  title={Judging llm-as-a-judge with mt-bench and chatbot arena},
  author={Zheng, Lianmin and Chiang, Wei-Lin and Sheng, Ying and Zhuang, Siyuan and Wu, Zhanghao and Zhuang, Yonghao and Lin, Zi and Li, Zhuohan and Li, Dacheng and Xing, Eric and others},
  journal={Advances in Neural Information Processing Systems},
  volume={36},
  year={2024}
}

@misc{luo2023latent,
      title={Latent Consistency Models: Synthesizing High-Resolution Images with Few-Step Inference}, 
      author={Simian Luo and Yiqin Tan and Longbo Huang and Jian Li and Hang Zhao},
      year={2023},
      eprint={2310.04378},
      archivePrefix={arXiv},
      primaryClass={cs.CV}
}

@article{luo2023lcm,
  title={LCM-LoRA: A Universal Stable-Diffusion Acceleration Module},
  author={Luo, Simian and Tan, Yiqin and Patil, Suraj and Gu, Daniel and von Platen, Patrick and Passos, Apolin{\'a}rio and Huang, Longbo and Li, Jian and Zhao, Hang},
  journal={arXiv preprint arXiv:2311.05556},
  year={2023}
}

@misc{chen2023pixartalpha,
      title={PixArt-$\alpha$: Fast Training of Diffusion Transformer for Photorealistic Text-to-Image Synthesis}, 
      author={Junsong Chen and Jincheng Yu and Chongjian Ge and Lewei Yao and Enze Xie and Yue Wu and Zhongdao Wang and James Kwok and Ping Luo and Huchuan Lu and Zhenguo Li},
      year={2023},
      eprint={2310.00426},
      archivePrefix={arXiv},
      primaryClass={cs.CV}
}

@misc{chen2024pixartdelta,
      title={PIXART-$\delta$: Fast and Controllable Image Generation with Latent Consistency Models}, 
      author={Junsong Chen and Yue Wu and Simian Luo and Enze Xie and Sayak Paul and Ping Luo and Hang Zhao and Zhenguo Li},
      year={2024},
      eprint={2401.05252},
      archivePrefix={arXiv},
      primaryClass={cs.CV}
}

@inproceedings{zhang2023adding,
  title={Adding conditional control to text-to-image diffusion models},
  author={Zhang, Lvmin and Rao, Anyi and Agrawala, Maneesh},
  booktitle={Proceedings of the IEEE/CVF International Conference on Computer Vision},
  pages={3836--3847},
  year={2023}
}

@article{xu2024hallucination,
  title={Hallucination is inevitable: An innate limitation of large language models},
  author={Xu, Ziwei and Jain, Sanjay and Kankanhalli, Mohan},
  journal={arXiv preprint arXiv:2401.11817},
  year={2024}
}

@article{byun2023transfer,
  title={Transfer Learning based Parameterized 3D Mesh Deformation with 2D Stylized Cartoon Character.},
  author={Byun, Sanghyun and Kim, Bumsoo and Shin, Wonseop and Jung, Yonghoon and Seo, Sanghyun},
  journal={KSII Transactions on Internet \& Information Systems},
  volume={17},
  number={11},
  year={2023}
}

@misc{MMPose_Contributors_OpenMMLab_Pose_Estimation_2020,
    author = {{MMPose Contributors}},
    title = {{OpenMMLab Pose Estimation Toolbox and Benchmark}},
    year = {2020},
    month = aug,
    howpublished = {\url{https://github.com/open-mmlab/mmpose}},
    note = {License: Apache-2.0}
}

@inproceedings{yang2011articulated,
  title={Articulated pose estimation with flexible mixtures-of-parts},
  author={Yang, Yi and Ramanan, Deva},
  booktitle={CVPR 2011},
  pages={1385--1392},
  year={2011},
  organization={IEEE}
}

@inproceedings{duan2019trb,
  title={Trb: a novel triplet representation for understanding 2d human body},
  author={Duan, Haodong and Lin, Kwan-Yee and Jin, Sheng and Liu, Wentao and Qian, Chen and Ouyang, Wanli},
  booktitle={Proceedings of the IEEE/CVF international conference on computer vision},
  pages={9479--9488},
  year={2019}
}

@article{muqeet2023video,
  title={Video Face Re-Aging: Toward Temporally Consistent Face Re-Aging},
  author={Muqeet, Abdul and Lee, Kyuchul and Kim, Bumsoo and Hong, Yohan and Lee, Hyungrae and Kim, Woonggon and Lee, KwangHee},
  journal={arXiv preprint arXiv:2311.11642},
  year={2023}
}

@inproceedings{bai2023ffhq,
  title={Ffhq-uv: Normalized facial uv-texture dataset for 3d face reconstruction},
  author={Bai, Haoran and Kang, Di and Zhang, Haoxian and Pan, Jinshan and Bao, Linchao},
  booktitle={Proceedings of the IEEE/CVF Conference on Computer Vision and Pattern Recognition},
  pages={362--371},
  year={2023}
}

@inproceedings{andriluka14cvpr,                author = {Mykhaylo Andriluka and Leonid Pishchulin and Peter Gehler and Schiele, Bernt},
 title = {2D Human Pose Estimation: New Benchmark and State of the Art Analysis},
booktitle = {IEEE Conference on Computer Vision and Pattern Recognition (CVPR)},                year = {2014},                month = {June}
}

@article{voleti2024sv3d,
  title={Sv3d: Novel multi-view synthesis and 3d generation from a single image using latent video diffusion},
  author={Voleti, Vikram and Yao, Chun-Han and Boss, Mark and Letts, Adam and Pankratz, David and Tochilkin, Dmitry and Laforte, Christian and Rombach, Robin and Jampani, Varun},
  journal={arXiv preprint arXiv:2403.12008},
  year={2024}
}

@article{reid2024gemini,
  title={Gemini 1.5: Unlocking multimodal understanding across millions of tokens of context},
  author={Reid, Machel and Savinov, Nikolay and Teplyashin, Denis and Lepikhin, Dmitry and Lillicrap, Timothy and Alayrac, Jean-baptiste and Soricut, Radu and Lazaridou, Angeliki and Firat, Orhan and Schrittwieser, Julian and others},
  journal={arXiv preprint arXiv:2403.05530},
  year={2024}
}

@article{brown2020language,
  title={Language models are few-shot learners},
  author={Brown, Tom and Mann, Benjamin and Ryder, Nick and Subbiah, Melanie and Kaplan, Jared D and Dhariwal, Prafulla and Neelakantan, Arvind and Shyam, Pranav and Sastry, Girish and Askell, Amanda and others},
  journal={Advances in neural information processing systems},
  volume={33},
  pages={1877--1901},
  year={2020}
}

@article{jing2023faithscore,
  title={Faithscore: Evaluating hallucinations in large vision-language models},
  author={Jing, Liqiang and Li, Ruosen and Chen, Yunmo and Jia, Mengzhao and Du, Xinya},
  journal={arXiv preprint arXiv:2311.01477},
  year={2023}
}

@inproceedings{lugmayr2022repaint,
  title={Repaint: Inpainting using denoising diffusion probabilistic models},
  author={Lugmayr, Andreas and Danelljan, Martin and Romero, Andres and Yu, Fisher and Timofte, Radu and Van Gool, Luc},
  booktitle={Proceedings of the IEEE/CVF conference on computer vision and pattern recognition},
  pages={11461--11471},
  year={2022}
}
}

\newpage
\appendix
\onecolumn

\section{Fewer Samples for PA-ICVL}  \label{supple:fewer_samples}

 {We demonstrate how the number of samples $N$ for PA-ICVL impacts on results. Including 5 samples for each class used in the main experiments, we gradually reduced the number of samples to 1 and 3. When we applied in-context learning, we saw that different values of $N$ resulted in different detection results, which is shown in Tab.} \ref{tab:N_samples}

\begin{table}[h]
    \centering
    \begin{tabular}{l|c|c|c} 
         \toprule
         Model (\textbf{D-5}) & $N=1$ & $N=3$ & $N=5$ \\
         \midrule
         GPT-4 Vision & 71\% & 73\% & 78\% \\
         Gemini 1.5 Pro & 72\% & 73\% & 80\% \\
         \bottomrule
    \end{tabular}
    \vspace{-0.5em}
    \caption{Results of final model according to the number of $N$.}
    \vspace{-1.0em}
    \label{tab:N_samples}
\end{table}

\section{Results on Image Transformation}

 {Based on the appearance of our dataset, it can be inferred that VLM may focus on the leg or arm regions. Thus, detecting hallucination might be easy in our settings. To address this conjecture, we separately evaluate the hallucination detection performance of two VLMs by flipping and rotating images. The evaluation results for character images flipped horizontally relative to the base model (D5) did not differ significantly. When evaluated using character images rotated $0.5\pi$, there was a significant difference in accuracy. The evaluation results are shown in Tab. 1.  This gap could mean that the VLM does not recognise rotated character images correctly, as opposed to forward facing characters. }

\vspace{-0.5em}
\begin{table}[h]
    \centering
    \begin{tabular}{l|c|c|c} 
         \toprule
         Model (\textbf{D-5}) & Base & Horizontal-Flip & $0.5\pi$ Rotation  \\
         \midrule
         GPT-4 Vision & 78\% & 76\% & 54\% \\
         Gemini 1.5 Pro & 80\% & 77\% & 61\% \\ 
         \bottomrule
    \end{tabular}
    \vspace{-0.5em}
    \caption{Results according to image transformation}
    \vspace{-1.0em}
    \label{table:rebuttal_pose_estimation}
\end{table}

\section{Visual Hallucination in Cartoon Domain} \label{suppl:cartoon_domain_hallucination}

\begin{figure}[ht]
  \centering
  \begin{subfigure}{0.45\linewidth}
    \includegraphics[width=\textwidth]{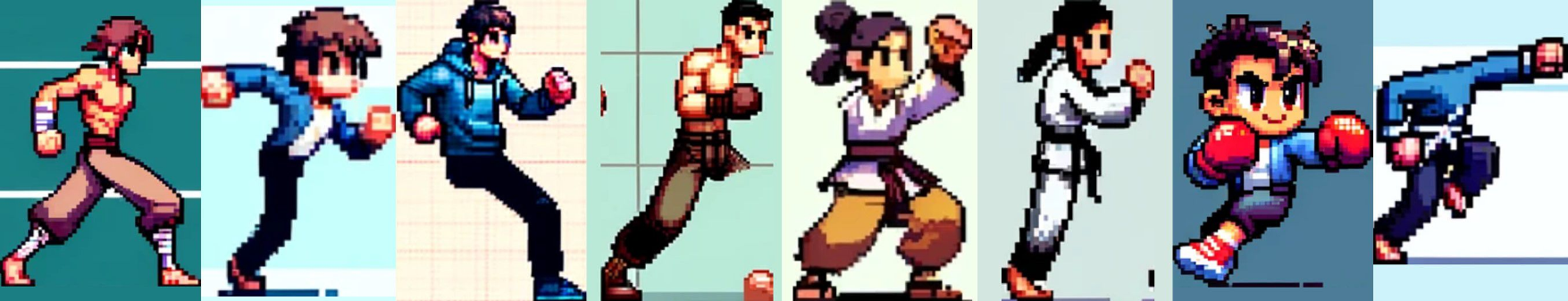}
    \caption{Case with \textit{few} body components.}
    \label{fig:few_type_hallucination}
  \end{subfigure}
  \quad  
  \begin{subfigure}{0.45\linewidth}
    \includegraphics[width=\textwidth]{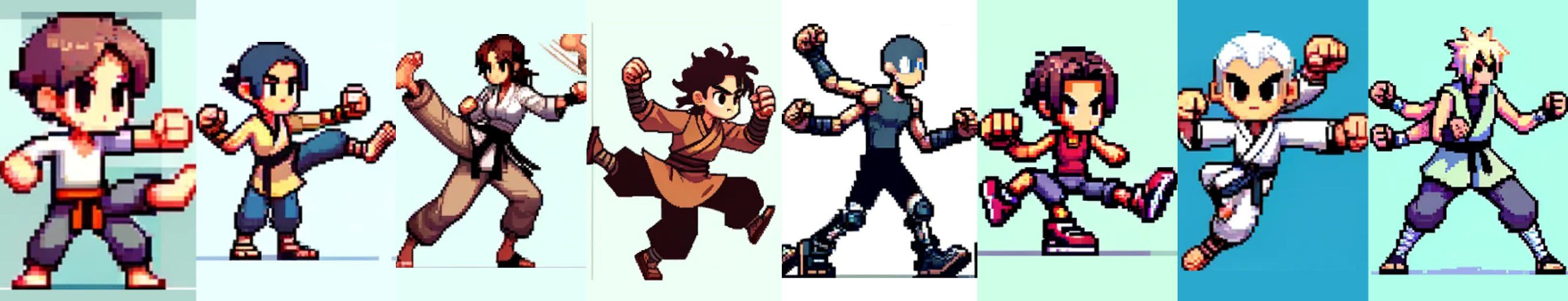}
    \caption{Case with \textit{many} body components.}
    \label{fig:many_type_hallucination}
  \end{subfigure}
  \caption{Hallucination classes in cartoon domain.}
  \label{fig:hallucination_class_in_cartoon}
\end{figure}

Cartoon domain has unique appearance. The cases and level of visual hallucination are quite different from realistic domain. When we generate many cartoon image from TTI, we found that there are two classes about hallucination tendency as shown in Fig. \ref{fig:hallucination_class_in_cartoon}: one is uncompleted whole-body having one arm, one leg or even no head as shown in Fig. \ref{fig:few_type_hallucination}, other one is over-depiction of body components such as three arms, three legs as shown in Fig. \ref{fig:many_type_hallucination}. These hallucination types led us utilize pose estimation for visual hallucination detection in cartoon image.

\begin{figure}[ht]
  \centering
  \begin{subfigure}[t]{0.12\linewidth}
    \centering
    \includegraphics[width=\linewidth]{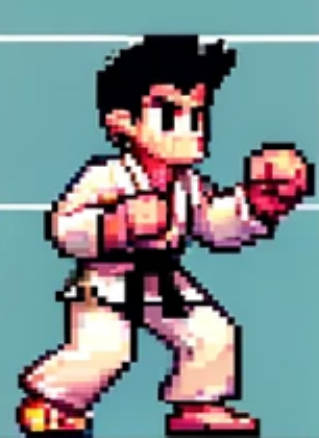}
    \caption{DALL-E3 \cite{betker2023improving}}
    \label{fig:dalle_3}
  \end{subfigure}\quad
  \begin{subfigure}[t]{0.12\linewidth}
    \centering
    \includegraphics[width=\linewidth]{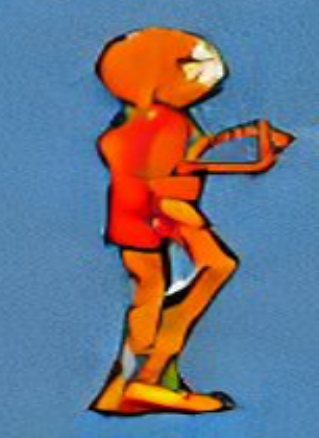}
    \caption{SDv2.1 \cite{luo2023latent}}
    \label{fig:sd_2_1}
  \end{subfigure}\quad
  \begin{subfigure}[t]{0.12\linewidth}
    \centering
    \includegraphics[width=\linewidth]{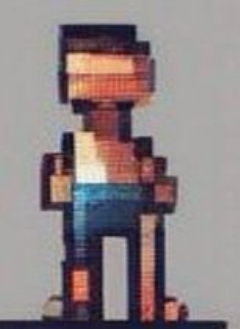}
    \caption{SDXLv1.0 \cite{podell2023sdxl}}
    \label{fig:sdxl_1_0}
  \end{subfigure}\quad
  \begin{subfigure}[t]{0.12\linewidth}
    \centering
    \includegraphics[width=\linewidth]{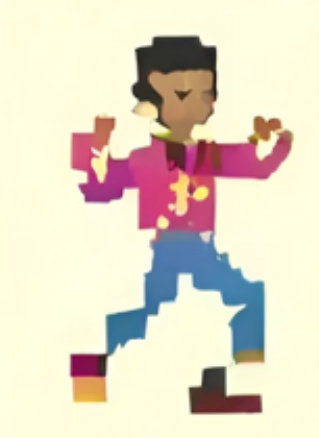}
    \caption{Karlo \cite{kakaobrain2022karlo-v1-alpha}}
    \label{fig:karlo}
  \end{subfigure}\quad
  \begin{subfigure}[t]{0.12\linewidth}
    \centering
    \includegraphics[width=\linewidth]{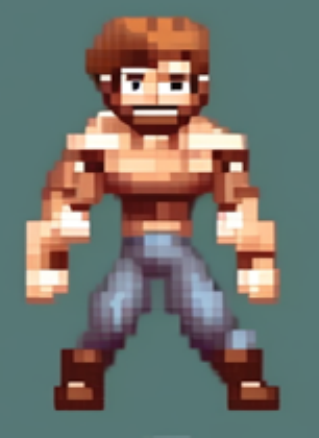}
    \caption{PIXART-$\alpha$ \cite{chen2023pixartalpha}}
    \label{fig:pix_art_alpha}
  \end{subfigure}\quad
  \begin{subfigure}[t]{0.12\linewidth}
    \centering
    \includegraphics[width=\linewidth]{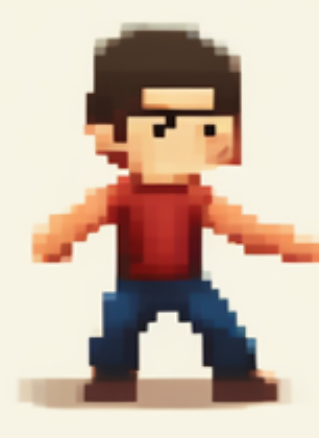}
    \caption{PIXART-$\alpha$ with LCM \cite{luo2023latent, luo2023lcm}}
    \label{fig:pix_art_alpha_lcm}
  \end{subfigure}
  \caption{Comparison for cartoon rendering quality with SOTA TTI models}
  \label{fig:comparison_tti}
\end{figure}

\section{Cartoon Rendering Comparison with Various Large TTI Models} \label{suppl:t2i_comaprison}

We evaluate TTI models which include DALL-E3 \cite{betker2023improving}, SDv2.1\footnote{https://huggingface.co/spaces/stabilityai/stable-diffusion} \cite{luo2023latent}, SDXLv1.0\footnote{https://huggingface.co/stabilityai/stable-diffusion-xl-base-1.0} \cite{podell2023sdxl}, Karlo\footnote{https://karlo.ai/} \cite{kakaobrain2022karlo-v1-alpha}, PIXART-$\alpha$\footnote{https://huggingface.co/spaces/PixArt-alpha/PixArt-alpha} \cite{chen2023pixartalpha}, PIXART-$\alpha$ with LCM\footnote{https://huggingface.co/spaces/PixArt-alpha/PixArt-LCM
}. For comparison in terms of cartoon rendering quality, we used same input prompt for every TTI models. For Karlo, the input prompts we used were also converted to combinations of words because, in Karlo website, we found that user typically used word format rather than sentences.

As shown in Fig. \ref{fig:comparison_tti}, we found that all the models generate non-plausible appearance for cartoon-pixel character except DALL-E3. Due to that, we used DALL-E3 for our TTI model.

\section{ChatGPT-4 Vision vs DALL-E3 API} \label{supple:dalle_api}

We found that there are some gaps about appearance tendency between ChatGPT-4 Vision through ChatGPT site\footnote{https://openai.com/gpt-4} and pure DALL-E3 API\footnote{https://openai.com/dall-e-3}. We conducted experiments by feeding the TTI prompts into both ChatGPT-4 Vision and the DALL-E3 API to generate images. Empirically, we observed that ChatGPT-4 Vision created images that were closer to our desired output with clean apparent structure compared to DALL-E3 API. We conjecture that this result is derived from ChatGPT's capability to refine and analyze the provided prompt, which in turn elevates its comprehension of the prompt, leading to the production of superior images.

\section{Non-Human-like Cartoon Character} \label{supple:non_human_like}

\begin{figure}[h]
\centering
    \includegraphics[width=0.6\linewidth]{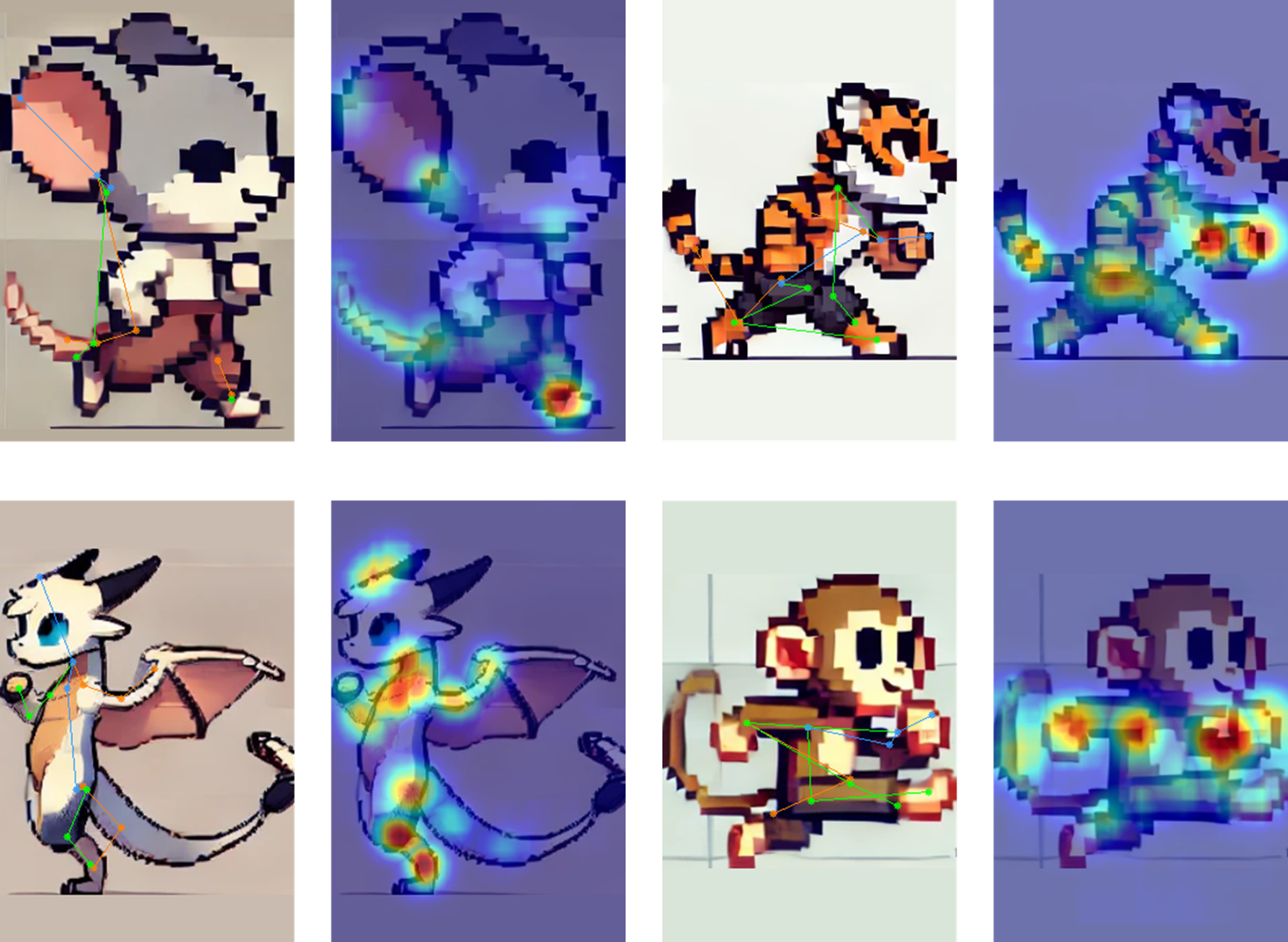 }
\caption{Failed pose estimation on non-human-like character.}	
\label{fig:non_human_like_pose}    
\end{figure}

 {We would like to extend the applicability or ours to non-human-like cartoon characters. To do so, we wanted to evaluate these samples by finding suitable input prompts for them. However, it was observed that the pose estimator was unable to extract accurate pose joints as shown in Fig. {\ref{fig:non_human_like_pose}}. Thus we cannot conduct evaluations with non-human-like character under the conclusion that pose guidance would not provide useful information to VLMs in PA-ICVL step. }

\section{Pose Estimator} \label{suppl:pose_estimator}

Here, we provide some details for our pose estimator including fine-tuning scheme and comparison on other off-the-shelf pose estimators.

\subsection{Comparison on Various Pose Estimators}

To use visually precise pose joint about cartoon character to VLM, we used our fine-tuned pose estimator (see Appendices \ref{suppl:fine_tuning} for details of fine-tuning). Here, we show the pose estimation performance with off-the-shelf pose estimators which include PoseAnything \cite{hirschorn2023pose}, OpenPose \cite{8765346}. For PoseAnythin, we used 1-Shot split-1 small model from official repository\footnote{https://github.com/orhir/PoseAnything}. For OpenPose, we used pose\_iter\_160000.caffemodel for MPII format from official repository\footnote{https://github.com/CMU-Perceptual-Computing-Lab/openpose}.

\begin{figure}[!ht]
  \centering
  \begin{subfigure}{0.17\linewidth}
    \centering
    \includegraphics[width=\linewidth]{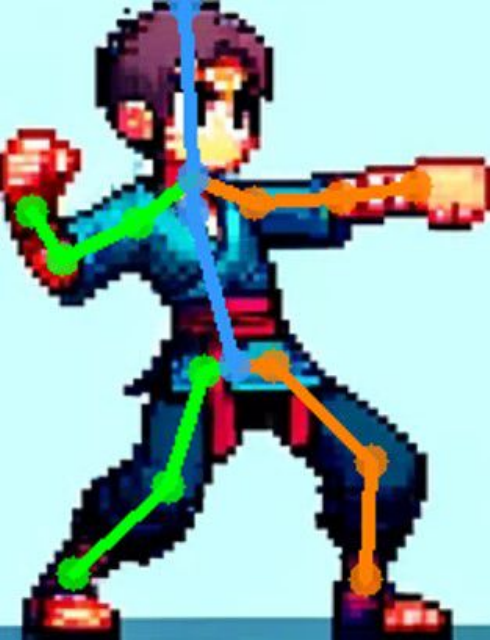}
    \caption{Our pose estimator}
    \label{fig:our_pose_estimator}
  \end{subfigure}\quad 
  \begin{subfigure}{0.17\linewidth}
    \centering
    \includegraphics[width=\linewidth]{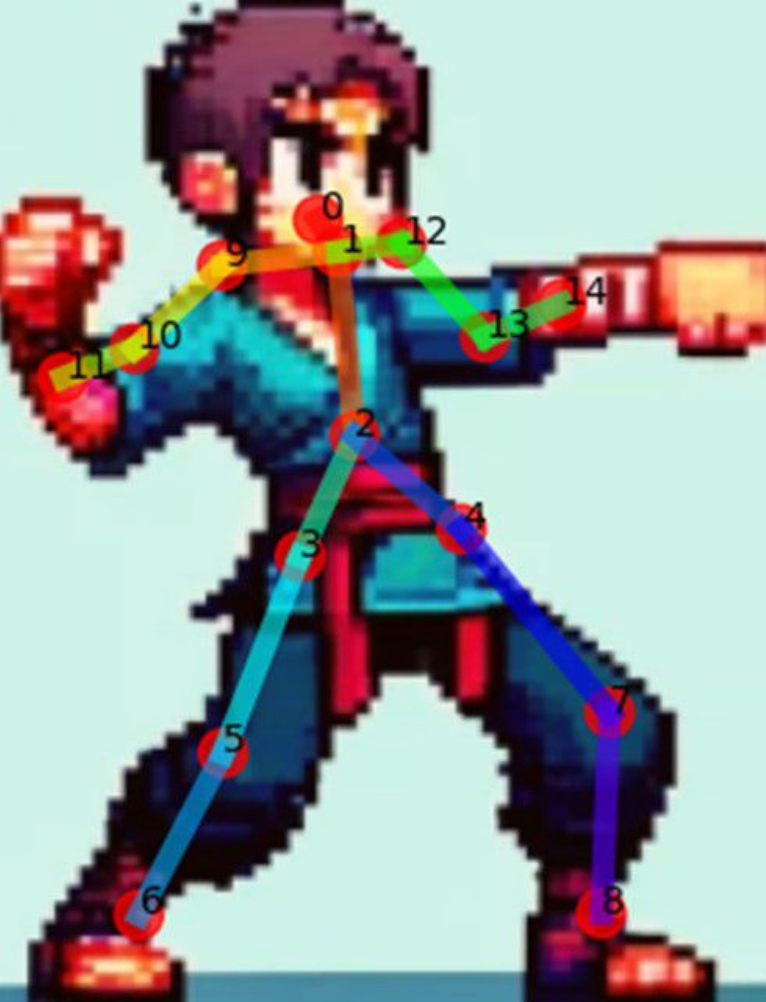}
    \caption{PoseAnything \cite{hirschorn2023pose}}
    \label{fig:pose_anything}
  \end{subfigure}\quad 
  \begin{subfigure}{0.17\linewidth}
    \centering
    \includegraphics[width=\linewidth]{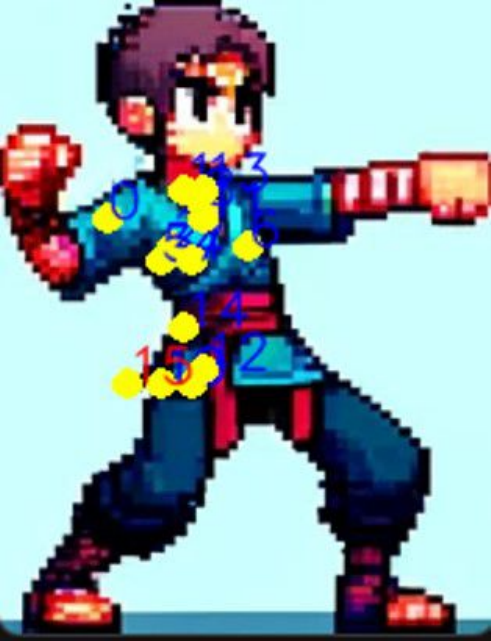}
    \caption{OpenPose \cite{8765346}}
    \label{fig:open_pose}
  \end{subfigure}
  \caption{Comparison of pose estimation.}
  \label{fig:comparison_pose_estimation}
\end{figure}

As shown in Fig. \ref{fig:comparison_pose_estimation}, we found that there is no one which can predict perceptually plausible joint. This is conjectured that cartoon domain distributed far away from pre-trained weights, leading requirement of fine-tuning with this domain.

\begin{figure}[h]
\centering
    \includegraphics[width=0.6\linewidth]{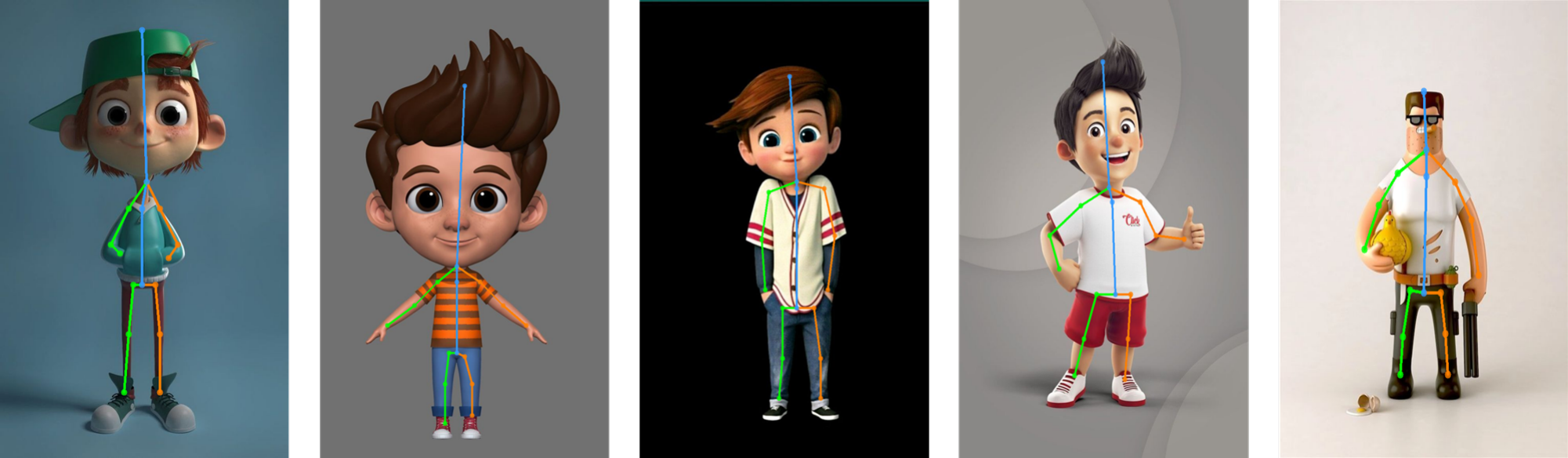}
\caption{Example of training dataset for pose estimation fine-tuning.}	
\vspace{-1.0em}
\label{fig:pose_estimation_dataset}    
\end{figure}

\subsection{Fine-tuning}
\label{suppl:fine_tuning}

We fine-tune the pose estimator based on HRNet-w48 \cite{sun2019deep} (Top-down arppoach) upon MMPose library \cite{MMPose_Contributors_OpenMMLab_Pose_Estimation_2020} with collected 3D cartoon dataset as shown in Fig. \ref{fig:pose_estimation_dataset}. Our training scheme is based on \cite{byun2023transfer}. For pose estimation settings, we used MPII-TRB \cite{duan2019trb} keypoint format (16 joint for whole-body) with inference image size as 384 by 256 using boundary padding. Including 2D illustration and rendered 2D images from 3D model shape, totalling 2400 images in animation, illustration and cartoon domains are used for fine-tuning for about 16K iterations with 32 batch size, achieving 0.8902 PCKh (Percentage of Correct Key-points head) \cite{yang2011articulated} at threshold 0.5.

\section{Used prompts} \label{suppl:input_prompt}
\subsection{TTI Input prompt list}
\noindent1. Please design a 2D motion frame pixel style character with a size of 256x384 pixels. Each action should be displayed on a separate row, with the first row being a \{kicking, punching, jumping, runnig, walking .. etc\} action (composed of 5 frames) and the second row being a \{kicking, punching, jumping, runnig, walking .. etc\} action (composed of 5 frames) that appears to be smoothly connected. The generated actions should not overlap, and the character s color should be simple. The entire sprite sheet should be 1792x1024 in size. \\

\noindent2. (Without detailed color \& overlap prompt) Please design a 2D pixel style character with a size of 256x384 pixels. Each action should be displayed on a separate row, with the first row being a \{kicking, punching, jumping, runnig, walking .. etc\} action (composed of 5 frames) and the second row being a \{kicking, punching, jumping, runnig, walking .. etc\} action (composed of 5 frames) that appears to be smoothly connected. The entire sprite sheet should be 1792x1024 in size. \\

\noindent3. (Without pixel size prompt) Please design a 2D motion frame pixel style character. Each action should be displayed on a separate row, with the first row being a {kicking, punching, jumping, runnig, walking .. etc\} action and the second row being a {kicking, punching, jumping, runnig, walking .. etc\} action that appears to be smoothly connected. The generated actions should not overlap, and the character s color should be simple. The entire sprite sheet should be 1792x1024 in size. 

\subsection{Instruction prompt list}
\noindent System Prompt \& Hallucination Definition : You are a hallucination detector, and your mission is to detect if the image has hallucinations.  Here I define hallucination as when a character is missing an arm, leg, or has an abnormal number of them (three legs, three arms .. etc). So you need to detect  the hallucination I defined in the image and visually describe the hallucination. So as a sample of how to detect this well, we'll provide the following prompts with a hallucination image, a normal image \{and joint image, joint file, heatmap image\}. \\

\noindent1. (\textbf{Model C}) Using RGB image - Correct class : This character is performing a \{kicking, punching, jumping, runnig, walking .. etc\} motion with an image of a correct human body with two arms and two legs.  This image of a correct human anatomy will be classified as \texttt{C} (correct class) in the future. Your task is to recognize images with correct human anatomy as \texttt{C} images.\\

\noindent2. (\textbf{Model C}) Using RGB image - Hallucination class : This character is performing a \{kicking, punching, jumping, runnig, walking .. etc\} motion with an abnormal character with \{three legs, three arms, no head, no arms, no legs, only one arm, only one leg\}. This image of abnormal human anatomy will be classified as \texttt{H} (hallucination class) in the future. Your task is to recognize images with abnormal human anatomy as \texttt{H} images.\\

\noindent3. (\textbf{Model D-1}) Using RGB \& Gaussian heatmap image - Correct class : This image is a pose heatmap obtained from the pose estimator. This character is performing a \{kicking, punching, jumping, runnig, walking .. etc\} motion with an image of a correct human body with two arms and two legs.  This image of a correct human anatomy will be classified as \texttt{C} (correct class) in the future. Your task is to recognize images with correct human anatomy as \texttt{C} images.\\

\noindent4. (\textbf{Model D-1}) Using RGB \& Gaussian heatmap image - Hallucination class : This image is a pose heatmap obtained from the pose estimator.  This character is performing a \{kicking, punching, jumping, runnig, walking .. etc\} with \{three legs, three arms, no head, no arms, no legs, only one arm, only one leg\}. This image of abnormal human anatomy will be classified as \texttt{H} (hallucination class)) in the future. Your task is to recognize images with abnormal human anatomy as \texttt{H} images.\\

\noindent5. (\textbf{Model D-2}) Using Overlapped heatmap image - Correct class : This image is a pose heatmap obtained from the pose estimator. This character is performing a \{kicking, punching, jumping, runnig, walking .. etc\} motion with an image of a correct human body with two arms and two legs.  This image of a correct human anatomy will be classified as \texttt{C} (correct class) in the future. Your task is to recognize images with correct human anatomy as \texttt{C} images.\\

\noindent6. (\textbf{Model D-2}) Using Overlapped heatmap image - Hallucination class : This image is a pose heatmap obtained from the pose estimator.  This character is performing a \{kicking, punching, jumping, runnig, walking .. etc\} with \{three legs, three arms, no head, no arms, no legs, only one arm, only one leg\}. This image of abnormal human anatomy will be classified as \texttt{H} (hallucination class)) in the future. Your task is to recognize images with abnormal human anatomy as \texttt{H} images.\\

\noindent7. (\textbf{Model D-3}) Using RGB \& overlapped heatmap image - Correct class : The first image is an RGB image of the character and the second image is a heatmap of the character's pose using the pose estimator. This character is performing a \{kicking, punching, jumping, runnig, walking .. etc\} motion with an image of a normal human body with two arms and two legs.  This image of a normal human anatomy will be classified as \texttt{C} (correct class) in the future. Your task is to recognize images with normal human anatomy as \texttt{C} images.\\

\noindent8. (\textbf{Model D-3}) Using RGB \& overlapped heatmap image - Hallucination class : The first image is an RGB image of the character and the second image is a heatmap of the character's pose using the pose estimator.  This character is performing a \{kicking, punching, jumping, runnig, walking .. etc\} motion with \{three legs, three arms, no head, no arms, no legs, only one arm, only one leg\}. This image of abnormal human anatomy will be classified as \texttt{H} (hallucination class) in the future. Your task is to recognize images with abnormal human anatomy as \texttt{H} images.\\

\noindent9. (\textbf{Model D-4}) Using RGB image \& Joint (image) - Correct class : The first image is an RGB image of the character and the second image is a keypoint of the character's pose using the pose estimator. This character is performing a  \{kicking, punching, jumping, runnig, walking .. etc\} motion with an image of a correct human body with two arms and two legs.  This image of a correct human anatomy will be classified as \texttt{C} (correct class) in the future. Your task is to recognize images with correct human anatomy as \texttt{C} images.\\

\noindent10. (\textbf{Model D-4}) Using RGB image \& Joint (image) - Hallucination class : The first image is an RGB image of the character and the second image is a keypoint of the character's pose using the pose estimator. This character is performing a  \{kicking, punching, jumping, runnig, walking .. etc\} motion with an abnormal character with \{three legs, three arms, no head, no arms, no legs, only one arm, only one leg\}. This image of abnormal human anatomy will be classified as \texttt{H} (hallucination class) in the future. Your task is to recognize images with abnormal human anatomy as \texttt{H} images.\\

\noindent11. (\textbf{Model D-5}) Using RGB image \& Joint (text) - Correct class : The first image is an RGB image of the character and the second file is a keypoint of the character's pose using the pose estimator. This character is performing a \{kicking, punching, jumping, runnig, walking .. etc\} motion with an image of a correct human body with two arms and two legs.  This image of a correct human anatomy will be classified as \texttt{C} (correct class) in the future. Your task is to recognize images with correct human anatomy as \texttt{C} images.\\

\noindent12. (\textbf{Model D-5}) Using RGB image \& Joint (text) - Hallucination class : The first image is an RGB image of the character and the second file is a keypoint of the character's pose using the pose estimator.   This character is performing a  \{kicking, punching, jumping, runnig, walking .. etc\} motion with an abnormal character with \{three legs, three arms, no head, no arms, no legs, only one arm, only one leg\}. This image of abnormal human anatomy will be classified as \texttt{H} (hallucination class) in the future. Your task is to recognize images with abnormal human anatomy as \texttt{H} images.\\
\end{document}